\documentclass{article}
\usepackage[round]{natbib}

\usepackage{arxiv}

\usepackage{booktabs} 
\usepackage{nicefrac}  
\usepackage{microtype}
\usepackage[utf8]{inputenc}

\usepackage[dvipsnames]{xcolor}
\usepackage{wrapfig}
\usepackage{multirow}

\usepackage{amssymb}
\usepackage{amsthm}
\usepackage{bbm}
\usepackage{array}
\usepackage[colorlinks=true,urlcolor=purple,
linkcolor=purple,citecolor=purple]{hyperref}
\usepackage{amsmath}
\usepackage{mathtools}
\usepackage{float}

\usepackage{mathrsfs}
\usepackage{graphicx}
\usepackage{color}

\usepackage{algorithm}
\usepackage{algpseudocode}
\usepackage{thmtools,thm-restate} %

\makeatletter
\def\paragraph{\@startsection{paragraph}{4}%
  \z@\z@{-\fontdimen2\font}%
  {\normalfont\bfseries}}
\makeatother

\usepackage{tikz}
\usetikzlibrary{decorations.markings, backgrounds, calc, positioning, shapes, shadows, arrows, fit, automata}
\usepackage{tikz-cd}
\usetikzlibrary{arrows.meta}
\tikzset{>={Latex}}

\newcommand{\frechet}{Fr\'{e}chet } %
\newcommand{\gwspec}{SpecGWL }

\newcommand{\R}{\mathbb{R}}

\newcounter{desccount}

\newcommand{\descref}[1]{\hyperref[#1]{#1}}

\newcommand{\dgw}{d_{\textrm{GW}}}		%
\newcommand{\dsgw}{d_{\textrm{GW}}^{\textrm{spec}}}	%

\newcommand{\tr}{\operatorname{tr}}

\DeclareMathOperator*{\argmin}{arg\,min}

\newtheorem{theorem}{Theorem}

\newtheorem{lemma}[theorem]{Lemma}

\makeatletter
\newcommand{\pushright}[1]{\ifmeasuring@#1\else\omit\hfill$\displaystyle#1$\fi\ignorespaces}
\newcommand{\pushleft}[1]{\ifmeasuring@#1\else\omit$\displaystyle#1$\hfill\fi\ignorespaces}
\makeatother

\title{Generalized Spectral Clustering via Gromov-Wasserstein Learning}

\author{
  Samir Chowdhury\\
  Stanford University\\
   \And
 Tom Needham \\
  Florida State University\\
}

\begin{document}
\maketitle

\begin{abstract}
  We establish a bridge between spectral clustering and Gromov-Wasserstein Learning (GWL), a recent optimal transport-based approach to graph partitioning. This connection both explains and improves upon the state-of-the-art performance of GWL. The Gromov-Wasserstein framework provides probabilistic correspondences between nodes of source and target graphs via a quadratic programming relaxation of the node matching problem. Our results utilize and connect the observations that the GW geometric structure remains valid for any rank-2 tensor, in particular the adjacency, distance, and various kernel matrices on graphs, and that the heat kernel outperforms the adjacency matrix in producing stable and informative node correspondences. Using the heat kernel in the GWL framework provides new multiscale graph comparisons without compromising theoretical guarantees, while immediately yielding improved empirical results. A key insight of the GWL framework toward graph partitioning was to compute GW correspondences from a source graph to a template graph with isolated, self-connected nodes. We show that when comparing against a two-node template graph using the heat kernel at the infinite time limit, the resulting partition agrees with the partition produced by the Fiedler vector. This in turn yields a new insight into the $k$-cut graph partitioning problem through the lens of optimal transport. Our experiments on a range of real-world networks achieve comparable results to, and in many cases outperform, the state-of-the-art achieved by GWL.
\end{abstract}

\section{INTRODUCTION}

The Gromov-Wasserstein (GW) problem is a nonconvex quadratic program whose solution yields the \emph{GW distance}, a variant of Wasserstein distance from classical optimal transport (OT) which is able to compare distributions defined on different metric spaces. This is accomplished by replacing classical Wasserstein loss with a loss function defined in terms of relational information coming from metric data. Because it is able to compare  distributions defined on a priori incomparable spaces, GW distance is increasingly finding applications for learning problems on irregular domains such as graphs \citep{hendrikson, netlsd,vayer2019optimal, xu2019gromov, xu2020gromov}. In this context, a graph can be considered as a metric space by endowing it with geodesic distance. A soft matching between nodes of two different graphs is obtained by choosing distributions on each graph's nodes (e.g., uniform distributions) and computing the GW optimal transport plan between them.

Applications of GW distance have been bolstered by the observation that the GW problem does not fundamentally require a \emph{metric} to operate \citep{pcs16}; i.e., the definition of the GW loss function extends to other forms of relational data \citep{gwnets}. \cite{xu2019scalable} used this observation to produce the state-of-the-art Scalable Gromov-Wasserstein Learning framework for graph matching and partitioning, which fundamentally uses the the adjacency matrix (as opposed to the shortest path distance matrix) of a graph. There are many ways to derive relational data from a graph beyond its distance and adjacency matrices, such as  its various graph Laplacians and their corresponding heat kernels. A limitation of the current literature is a lack of  tools for ascertaining if the adjacency matrix (or any other rank-2 tensor derived from a graph) is optimal in some sense beyond empirical benchmarks. Thus the potential flexibility of the GW framework for graph analysis remains largely unexplored.

Here we study the GW graph OT problem by representing graphs via heat kernels rather than adjacency matrices. This amounts to finding soft correspondences between the nodes of two graphs by comparing spectral, rather than adjacency, data. We refer to this as the \gwspec framework, reserving GWL for the adjacency-based framework of \cite{xu2019scalable}. 

Numerical experiments\footnote{\url{https://github.com/trneedham/Spectral-Gromov-Wasserstein}} demonstrate that \gwspec outperforms GWL in graph partitioning tasks. Moreover, a main goal of this paper is to introduce tools for studying the GW problem in a more rigorous manner. We use a Markov Chain sampling technique to explore the energy landscape of the GW loss function for adjacency and heat kernel graph representations, showing empirically that \gwspec loss has fewer spurious local minima and a 10x acceleration in convergence of gradient descent over GWL loss. We also introduce a visualization technique which allows one to ascertain the quality of soft node matchings obtained by any GW method. Examples of this technique intuitively demonstrate the idea that heat kernel-based matchings more faithfully preserve global graph structure than adjacency-based matchings. Finally, we establish theoretical results on the sparsity of optimal couplings in the \gwspec framework and on the precise relation of \gwspec graph partitioning to classical spectral clustering. The latter result creates a novel connection between spectral clustering and optimal transport.

\paragraph{Related literature.}
Gromov-Wasserstein distance---as used in current ML settings---was introduced by \cite{dgh-sm} as a convex relaxation scheme for the Gromov-Hausdorff distance between different metric spaces. Further theoretical development was carried out by \cite{dghlp-focm,sturm2012space}, and a related formulation was introduced by \cite{sturm2006geometry} to study the convergence of sequences of metric measure spaces. The idea of using heat kernels for GW matching goes back to the \emph{Spectral Gromov-Wasserstein distance} introduced by \cite{memoli2011spectral} in the context of Riemannian manifolds to further develop the notion of \emph{Shape-DNA} introduced by \cite{reuter2006laplace}. A related theoretical construction also appeared in \cite{kasue1994spectral}. The Riemannian heat kernel has been celebrated for its \emph{multiscale} and \emph{informative} properties \citep{sun2009concise}---the former refers to the observation that the heat kernel defines a family of Gaussian filters that get progressively shorter and wider as $t\to \infty$, and the latter refers to the classical lemma of Varadhan showing that at the small time limit, the log-heat kernel approximates the geodesic distance on a Riemannian manifold. A surprising result due to \cite{sun2009concise} in this direction is that under mild conditions, the collection of traces of the heat kernel forms an isometry invariant of a manifold despite giving up most of the information contained in the heat kernel. Heat kernel traces have recently been used for graph comparison by \cite{netlsd,tsitsulin2020shape}, who showed that the desirable properties of the heat kernel for Riemannian manifolds have natural and informative analogues in the setting of graphs. Heat kernels have been incorporated into GW-based graph matching in recent work of \cite{barbe2020graph}, where they are used to augment the matching process in the Fused GW framework of \cite{vayer2018fused}. Our work is distinguished from that of \cite{barbe2020graph} in that we use heat kernels directly to define a GW loss function for graph matching, whereas \cite{barbe2020graph} employs heat kernels to improve feature space matchings for attributed graphs. There is a deep literature on spectral graph comparison, and a few other references include works of \cite{patro2012global, bronstein2013heat, hu2014stable, nassar2018low, dong2020copt}.

Starting from its early applications in computer vision \citep{dgh-sm, dghlp-focm, schmitzer2013modelling}, GW distance has been applied to alignment of word embedding spaces \citep{alvarez2018gromov}, learning of generative models across different domains \citep{bunne2019learning}, graph factorization \citep{xu2019gromov}, and learning autoencoders \citep{xu2020learning}. Related theoretical directions include the \emph{sliced GW} of \cite{vayer2019sliced} and the \emph{Gromov-Monge} problem studied by \cite{memoli2018gromov}. \cite{sturm2012space} studied the structure of geodesics and gradient flows in GW geometry, and recently these techniques were utilized by \cite{gwa} to create a Riemannian framework for performing averaging and tangent PCA across different graph-structured domains. Unbalanced formulations of the GW problem permitting input probability measures of different total sums have been studied by \cite{chapel2020partial, de2020entropy,sejourne2020unbalanced}.

\section{SPECTRAL GW DISTANCES}

\paragraph{Gromov-Wasserstein Distance.} 
The following can be stated for Borel probability measures on Polish spaces; however, we are interested in the finite setting where measure-theoretic complications do not arise, so we freely use matrix-vector notation. Given probability distributions $p,q$ on finite sets $X,Y$, a coupling of $p$ and $q$ is a joint probability measure $C$ on $X \times Y$ with marginals $p$ and $q$; i.e., $C = (C_{ij}) \in \R^{|X| \times |Y|}$ satisfies equality constraints $C1^{|Y|\times 1} = p$ and $C^T 1^{|X|\times 1} =q$ ($1^{m \times n}$ denoting the $m\times n$ matrix of all ones), and entrywise inequality constraints $0 \preceq C \preceq 1$. The set of all couplings of $p$ and $q$ is denoted $\mathcal{C}(p,q)$---this is a convex polytope in $\R^{|X| \times |Y|}$. Given functions $f^X:X\times X \to \R$, $f^Y:Y\times Y \to \R$, written as square matrices $F^X, F^Y$, the \emph{GW problem} solves:  
\begin{equation}
    \min_{C \in \mathcal{C}(p,q)}  \sum_{i,k} \sum_{j,l} ( F^X_{ik} - F^Y_{jl} )^2 C_{ij}C_{kl}.\label{eq:gwloss}
\end{equation}
The GW problem is a nonconvex quadratic program over a convex domain for which approximate solutions may be obtained via projected gradient descent \citep{pcs16} or Sinkhorn iterations with an entropy \citep{s16} or KL divergence regularizer \citep{xu2019gromov}. Computational implementations can be found in the Python Optimal Transport library of \cite{pot}.

Historically, the GW problem was defined by \cite{dgh-sm} in the setting of \emph{metric measure (mm) spaces}, where its solution leads to a metric known as the Gromov-Wasserstein distance. A finite mm space $(X,d,p)$ consists of a finite set of points $X$, a metric function $d$ written as a matrix $d = (d_{ik}) \in \R^{|X|\times |X|}$, and a probability distribution $p$ on $X$ written as a vector $p = (p_i) \in [0,1]^{|X|\times 1}$. Given mm spaces $(X,d^X,p), (Y,d^Y,q)$, the Gromov-Wasserstein distance is defined as:
\begin{equation} 
\dgw(X,Y)^2 := \min_{C \in \mathcal{C}(p,q)}  \sum_{i,k} \sum_{j,l} ( d^X_{ik} - d^Y_{jl} )^2 C_{ij}C_{kl}. \label{eq:gwdist}
\end{equation}
\cite{pcs16} observed that solving the GW problem \eqref{eq:gwloss} with input matrices that are not strictly distances---in particular kernel matrices---still leads to a discrepancy that is informative when comparing matrices of different sizes and arising in incompatible domains. This idea was pushed further by the theoretical work of \cite{gwnets}, which showed that \emph{any} square matrix representation of a graph, including the adjacency matrix, can be used in the GW problem \eqref{eq:gwloss} to obtain a bona fide distance (actually a pseudometric). This observation was used by \cite{xu2019gromov, xu2019scalable} to create a unified \emph{Gromov-Wasserstein Learning} (GWL) framework for unsupervised learning tasks on graphs---e.g. finding node correspondences between unlabeled graphs and graph partitioning---with state-of-the-art performance. We now formulate the GWL framework precisely.

\paragraph{Gromov-Wasserstein Learning on Graphs.} 

Let $G=(V,E)$ be a finite, unweighted, possibly directed graph. We refer to $V$ as the set of nodes and $E$ as the set of edges. Let $A:V\times V \to \{0,1\}$ and $D:V\to \mathbb{Z}$ denote the \emph{adjacency} and \emph{degree} functions defined as $A(v,w) := 1$ if $(v,w) \in E$, 0 otherwise, and $D(v) := |\{w : (v,w) \in E\}|$.
Given an ordering on $V$, these functions can be represented as matrices in $\R^{|V| \times |V|}$, and we will switch between both the function and matrix forms. In addition to the adjacency function, which fully encodes a graph, there are numerous \emph{derived representations} which capture information about a graph. An example is the geodesic distance function $d$ which contains all the shortest path lengths in the graph. In particular, the graph geodesic distance representation was used by \cite{hendrikson} to solve graph matching problems via the original mm-space GW distance formulation \eqref{eq:gwdist} of \cite{dgh-sm}.

One of the key ideas in the GWL framework is to use \emph{adjacency matrices} in the GW problem. Specifically, let $G$ and $H$ be graphs with distributions $p$ and $q$ on their nodes and let $A^G$ and $A^H$ denote their adjacency matrices. The GWL framework considers the loss 
\[
\mathrm{Adj} = \mathrm{Adj}_{G,p,H,q}:\mathcal{C}(p,q) \rightarrow \R
\]
defined by
\begin{equation}\label{eqn:gwl_loss}
    \mathrm{Adj}(C) = \sum_{i,k} \sum_{j,l} (A^G_{ik} - A^H_{jl})^2 C_{ij}C_{kl},
\end{equation}
which we refer to as \emph{adjacency loss}. The minimum of $\mathrm{Adj}(C)^{1/2}$ defines a distance between the pairs $(G,p)$ and $(H,q)$---we refer to such pairs as \emph{measure graphs} and assume for convenience that distributions are fully supported. If $p$ and $q$ are themselves derived from adjacency data then the minimizer $C$ of \eqref{eqn:gwl_loss} provides a natural soft correspondence between the nodes of $G$ and $H$. For example, \cite{xu2019scalable} consider the family of node distributions $p=(p_1,\ldots,p_n)^T$, where
\begin{equation}\label{eqn:distributions}
    p_j = \frac{\overline{p}_j}{\sum_{k=1}^n \overline{p}_k}, \qquad \overline{p}_j = (\mathrm{deg}(v_j) + a)^b,
\end{equation}
where $p_j = p(v_j)$ for $v_j$ a vertex of $G$, $a \geq 0$ is used to enforce the full support condition and the exponent $b \in [0,1]$ allows interpolation between the \emph{uniform distribution} and the \emph{degree distribution}.

\paragraph{Heat Kernels.} 
Our contributions begin by studying the structure imposed on the GW problem \eqref{eq:gwloss} by \emph{spectral losses} that we describe next.
Given an undirected graph $G=(V,E)$, let $L^2(V)$ denote the linear space of functions $f:V \to \R$. The \emph{Laplacian} of $G$ is the operator $L:L^2(V) \to L^2(V)$ defined by
\[ L(\phi)(v) := D(v)\phi(v) - \sum_{(v,w) \in E} \phi(w).\]
After fixing an ordering on $V$, we can use matrix-vector notation to write $\R^{|V|\times |V|} \ni L=D-A$ and $\phi \in \R^{|V|\times 1}$. The Laplacian is symmetric positive semidefinite, so the spectral theorem guarantees an eigendecomposition $L=\Phi\Lambda\Phi^T$ with real, nonnegative eigenvalues $\lambda_1 \leq \lambda_2 \leq \ldots \leq \lambda_n$ arranged on the diagonal of $\Lambda$. The corresponding eigenvectors $\phi_1,\phi_2,\ldots, \phi_n$ are arranged as the columns of $\Phi$. There are several variants of the Laplacian with analogous properties, including the \emph{normalized Laplacian} given by $I - D^{-{1/2}}AD^{-{1/2}}$. This is the version that we will typically use for experiments.

For a (strongly connected) directed graph $G=(V,E)$, we use the normalized Laplacian defined by \cite{chung2005laplacians} via the language of random walks. Consider the transition probability matrix $P$ defined by writing $P_{ij} = 1/D_i$ if $(i,j) \in E$, 0 otherwise. By Perron-Frobenius theory, there is a unique left eigenvector $\psi$ with all entries positive such that $\psi^TP = \psi^T$. The directed graph Laplacian is defined as $L:= I -(\Psi^{1/2}P\Psi^{-1/2} + \Psi^{-1/2}P^T\Psi^{1/2})/2$, where $\Psi = \mathrm{diag}(\psi)$.

In what follows, we use $L$ generically to refer to any of the Laplacians defined above. The \emph{heat equation} on a graph $G$ (either undirected or directed) is then given as $du/dt = -Lu$, where $u \in L^2(V \times \R_{>0})$. If $u$ represents a time-dependent heat distribution on the nodes of $G$, the heat equation describes heat diffusion according to Newton's Law. The \emph{heat kernel} is the fundamental solution to this heat equation, given in closed form as $K^t = \exp(-tL) = \Phi\exp(-t\Lambda)\Phi^T$.

\paragraph{Spectral GW Distance.} 
Given measure graphs $(G,p)$ and $(H,q)$, we consider the \emph{spectral loss}
\[
\mathrm{Spec}^t = \mathrm{Spec}^t_{G,p,H,q}:\mathcal{C}(p,q) \rightarrow \R
\]
defined by
\begin{equation}\label{eqn:gwspec_loss}         \mathrm{Spec}^t(C) = \sum_{i,k} \sum_{j,l} (K^{G,t}_{ik} - K^{H,t}_{jl})^2 C_{ij}C_{kl}
\end{equation}
for each $t > 0$, where $K^{G,t}$ and $K^{H,t}$ are the heat kernels of $G$ and $H$, written in matrix form. We then obtain a one parameter family of pseudometrics
\begin{equation}\label{eqn:spec_gw_dist}
\dsgw[t]((G,p),(H,q)) := \min_{C \in \mathcal{C}(p,q)} \mathrm{Spec}^t(C)^{1/2}
\end{equation}
on the space of measure graphs. As in the GWL framework, choosing node distributions $p$ and $q$ from the family \eqref{eqn:distributions}  yields minimizing couplings $C$ which give meaningful correspondences between the nodes of $G$ and $H$. As $t$ varies, one obtains multiscale couplings between nodes, with small $t$ encoding local and large $t$ encoding global structure (this is made precise in Section \ref{sec:graph_partitioning}). Intuitively, the heat kernel replaces each node of a graph with a Gaussian filter with width controlled by $t$, progressively ``smoothing'' the graph. Different smoothing levels emphasize multiscale geometric and topological features that drive coupling optimization; Figure \ref{fig:hkscale_tsne_composite} in the Supplementary Materials shows clusters in coupling space corresponding to matchings at different scales. The ``best'' choice of $t$ emphasizes the feature scale which is most important for a given task.

One can further define an overall pseudometric by, say, considering the function 
\[
f_{G,p,H,q}(t) = \dsgw[t]((G,p),(H,q))
\]
and defining $\dsgw := \|\xi \cdot f_{G,p,H,q}\|$ for an appropriately chosen norm on functions $\R \rightarrow \R$ and normalizing function $\xi$ preventing blow up of the norm; for example, taking $\xi(t) = \exp(-(t + t^{-1}))$ and the $\ell_\infty$ norm here is analogous to the spectral GW distance between Riemannian manifolds defined by \cite{memoli2011spectral}. This is an interesting direction of  future research, but we found it most useful in our applications to instead treat $t$ as a scale parameter that is tuned via cross-validation.

\paragraph{Properties of \gwspec\!\!\!.} For measure graphs $(G,p)$ and $(H,q)$, fix $t> 0$ and write $J:= K^{G,t}, K:= K^{H,t}$. After expanding the square and invoking the marginalization constraints (\cite{s16} provide an explicit derivation), one sees that minimizing \eqref{eqn:gwspec_loss} is equivalent to maximizing $\langle JC, CK\rangle$ subject to $C \in \mathcal{C}(p,q)$, where $\langle \cdot, \cdot \rangle$ denotes the Frobenius inner product. Because the graph heat kernel is symmetric positive definite, we take Cholesky decompositions $J=U^TU$, $K=V^TV$ to write 
\begin{align*}
    \langle JC,CK \rangle &= \tr((JC)^T CK) = \tr(C^T U^T U C V^T V) \\
    &= \tr(V C^T U^T U C V^T) = \|UCV^T\|^2.
\end{align*}
The map $C\mapsto \|UCV^T\|^2$ is convex. We record this as the following lemma, which appeared previously as \cite[Lemma 4.3]{alvarez2019towards}.

\begin{lemma}
\label{lem:concave}
For each $t > 0$, spectral loss \eqref{eqn:gwspec_loss} is minimized over the convex polytope $\mathcal{C}(p,q)$ by a maximizer of the convex function $C\mapsto \langle K^{G,t}C,CK^{H,t}\rangle$.
\end{lemma}

Optimization problems of this type are not tractable to solve deterministically, but we demonstrate experimentally that approximation via gradient descent enjoys faster convergence and fewer spurious local minima than adjacency loss \eqref{eqn:gwl_loss}. 
It has been empirically observed by \cite{xu2019gromov} that (local) minimizers of adjacency loss tend to become sparse, and this sparsity is crucial in the gradient descent-based algorithm of \cite{gwa} for computing averages of networks. 
A key advantage of our spectral setting \eqref{eqn:gwspec_loss} over \eqref{eq:gwdist} or \eqref{eqn:gwl_loss} is that this empirical observation of sparsity admits a formal proof.

\begin{theorem}
\label{thm:sparse-coup}
Let $(G,p)$ and $(H,q)$ be measure graphs on $\sim n$ nodes. Then for any $t > 0$, there is a minimizer of spectral loss with $o(n)$ nonzero entries.
\end{theorem}

\paragraph{Complexity of \gwspec\!\!\!.} Assuming graphs of comparable size $n$, the eigendecomposition incurs a time complexity of $O(n^3)$ and memory complexity of $\Theta(n^2)$. Computing the GW loss using gradient descent involves computing $\nabla \langle JC,CK \rangle = J^TCK + JCK^T$, which incurs a time complexity of $O(n^3\log(n))$ \citep{kolouri2017optimal} and memory complexity of $O(n^2)$. Regularized methods with Sinkhorn iterations still require paying a cost for matrix multiplication \citep{pcs16}. However, accelerations have already been proposed: \cite{netlsd} suggest methods for approximating heat kernels in $O(n^2)$ operations and \cite{xu2019scalable} propose a recursive divide-and-conquer approach to reduce the complexity of the GW comparison to $O(n^2\log n)$. Note that because heat kernel matrices are dense, we lose the advantages of sparse matrix operations.

\section{GRAPH PARTITIONING}\label{sec:graph_partitioning}

\paragraph{Graph Partitioning Method.}\label{subsec:graph_partitioning_method}

Graph partitioning is a crucial unsupervised learning task used for community detection in social and biological networks \citep{girvan2002community}. The goal is to partition the vertices of a graph into some number of clusters $m$ in accordance with the \emph{maximum modularity principle}---edges within clusters are dense, while edges between clusters are sparse. \cite{xu2019scalable} proposed a GW-based approach to graph partitioning where an $m$-way partition of a measured graph $(G,p)$ is obtained by minimizing the following variant of adjacency loss \eqref{eqn:gwl_loss}:
\begin{equation}\label{eqn:gwl_partition_loss}
    \mathcal{C}(p,q) \ni C \mapsto \sum_{i,k} \sum_{j,l} (A_{ik} - Q_{jl})^2 C_{ij} C_{kl},
\end{equation}
where $A$ is the adjacency matrix of $G$, $Q = \mathrm{diag}(q)$ and $q$ is a distribution estimated by sorting the weights of $p$, sampling $m$ values via linear interpolation and renormalizing. Intuitively, $Q$ is the weighted adjacency matrix of a graph on $m$ nodes with only self-loops---an \emph{ideally clustered template} graph. The heuristic for choosing the distribution $q$ in this manner is that if within-cluster nodes of the graph $G$ have similar degrees then this method allows a node in $Q$ to accept all of the mass from this cluster. A minimizer $C$ of \eqref{eqn:gwl_partition_loss} defines an $m$-way partition of $G$: each node $v_i$ of $G$ is assigned a label in $\{1,\ldots, m\}$ according to the column index of the maximum weight in row $i$ of $C$.

Soft-matching nodes of the target graph to an ideally clustered template is intuitively appealing and it was shown by \cite{xu2019gromov} that this method achieves state-of-the-art performance. We propose a variant of the algorithm: letting $K^t$ denote the heat kernel for $G$ at some $t > 0$, we minimize
\begin{equation}\label{eqn:spec_gwl_partition_loss}
    \mathcal{C}(p,q) \ni C \mapsto \sum_{i,k} \sum_{j,l} (K^t_{ik} - Q_{jl})^2 C_{ij} C_{kl},
\end{equation}
with $Q$ defined as above. For each $t > 0$, we obtain an optimal coupling which is used to partition $G$ as described above. Experimental results in Section \ref{subsec:graph_partitioning} show that this change to the algorithm gives a significant performance boost over the adjacency-based version.

\paragraph{Connection to Spectral Clustering.}

Let $G$ be an undirected, connected graph with graph Laplacian $L$. The connectivity of $G$ implies that $L$ has exactly one zero eigenvalue with  constant eigenvector. Assume for simplicity that the multiplicity of the smallest positive eigenvalue of $L$ is one. A fundamental concept in spectral graph theory is that the corresponding eigenvector---the \emph{Fiedler vector} of $G$---gives a 2-way partitioning of $G$ with good theoretical properties: nodes of $G$ are partitioned according to the sign of their entry in the Fiedler vector \citep{fiedler1973algebraic}. We refer to this as the \emph{Fiedler partition of $G$}, and find that it arises as a special case of spectral GW partitioning:

\begin{theorem}\label{thm:fiedler_partition}
Let $G$ be a connected graph whose first positive eigenvalue has multiplicity one, endowed with the uniform node probability distribution $p$. For sufficiently large $t$, the $2$-way partition of $G$ derived from a minimizer of \eqref{eqn:spec_gwl_partition_loss} agrees with the Fiedler partitioning.
\end{theorem}

The theorem demonstrates a novel connection between optimal transport and classical spectral clustering and shows that partitioning graphs via spectral GW matching \eqref{eqn:spec_gwl_partition_loss} is a generalization of these classical methods. For the small-$t$ regime, note that the heat kernel of $G$ has Taylor expansion $K^t = I_n + tL + O(t^2)$, where $I_n$ is the $n \times n$ identity matrix. Thus for low values of $t$, spectral GW partitioning is driven by matchings of graph Laplacians, which contain local adjacency information.

\section{EXPERIMENTS}

We present several numerical experiments demonstrating the boost in performance obtained by using heat kernels in the GW problem rather than adjacency matrices. In experiments with undirected graphs, we used the normalized graph Laplacian to construct heat kernels---results were qualitatively similar using heat kernels of the standard Laplacian, but we found some boost in quantitative performance in graph partitioning when using the normalized version. Experiments with directed graphs use Chung's normalized Laplacian.

\paragraph{Energy Landscapes and Convergence Rates.}\label{subsec:energy_landscapes}

Adjacency loss \eqref{eqn:gwl_loss} is highly nonconvex, while Lemma \ref{lem:concave} shows that minimizing spectral loss \eqref{eqn:gwspec_loss} is equivalent to maximizing a convex function over a convex polytope. While the latter optimization problem is still intractable, we now demonstrate that its approximation via gradient descent is well-behaved.
In each trial, two random graphs from the \textbf{IMDB-Binary} \citep{yanardag2015deep} actor collaboration graph  dataset (1000 graphs with 19.77 nodes and 96.53 edges on average) are selected. The nodes of the graphs are endowed with uniform distributions $p$ and $q$ (results obtained when using other distributions from the family \eqref{eqn:distributions} are similar). An ensemble of couplings between these measures is generated by running a custom Markov Chain Monte Carlo (MCMC) hit-and-run sampler \citep{smith1984efficient} on the coupling polytope $\mathcal{C}(p,q)$ (details in Supplementary Materials). We sampled 100 points in the polytope by running 100,000 MCMC steps and subsampling uniformly. Using each coupling in the ensemble as an initialization, we run projected gradient descent on adjacency loss \eqref{eqn:gwl_loss} and spectral loss \eqref{eqn:gwspec_loss} with $t=5$, $10$ and $20$ and record the loss at the local minimum from each initialization. This process is repeated 100 times (100 choices of pairs of graphs).

Statistics for the experiment are reported in Table \ref{tab:mean_stats_energy_landscapes}. For each method, we report the mean time for convergence of each gradient descent. For each trial, we obtain a distribution of losses from the 100 initializations in the ensemble, with the minimum loss treated as the putative global minimum. The ``Worst Error'' for each trial is $(\mbox{max loss}- \mbox{min loss})/\mbox{min loss}$. We report the mean Worst Error over 100 trials. In available packages, gradient descent for GW matching is by default initialized with the product coupling $C = pq^T$, so we also report the ``Product Error'' $(\mbox{product loss}- \mbox{min loss})/\mbox{min loss}$, averaged over all trials. We note that the absolute losses of the adjacency and heat kernel methods are not directly comparable, which is why we report these relative errors. The heat kernel representations provide an order of magnitude speed up of convergence, with decreasing error as the $t$ parameter increases; e.g., for $t=20$, all initializations converge to a coupling with less than 0.1\% error. More details are provided in the Supplementary Materials.

\begin{table}[!t]
\caption{Results of Energy Landscape Experiment.}
\label{tab:mean_stats_energy_landscapes}
\begin{center}
\begin{tabular}{l c c c}
\toprule
Loss & Time (s) & Err. (\%) & Prod. Err. (\%) \\\midrule
$\mathrm{Adj}$ & .0230 & 24.21 & 4.81 \\
$\mathrm{Spec}^5$ & .0014 & 22.35 & 6.00 \\
$\mathrm{Spec}^{10}$ & .0013 & 3.86 & 1.34 \\
$\mathrm{Spec}^{20}$ & .0010 & 0.05 & .02 \\
\bottomrule
\end{tabular}
\end{center}
\end{table}

\begin{figure*}
\centering
\includegraphics[width = 0.45\textwidth]{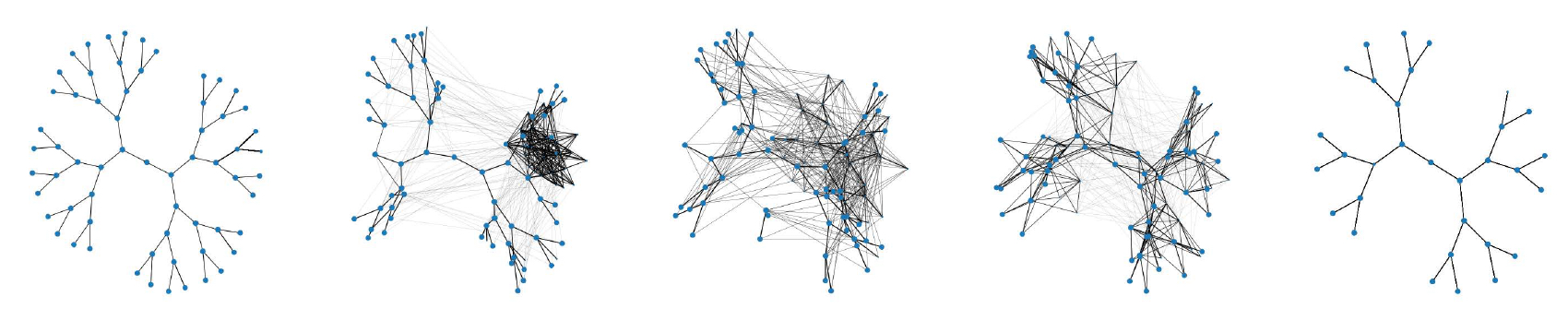}
\hspace{.2in}
\includegraphics[width = 0.45\textwidth]{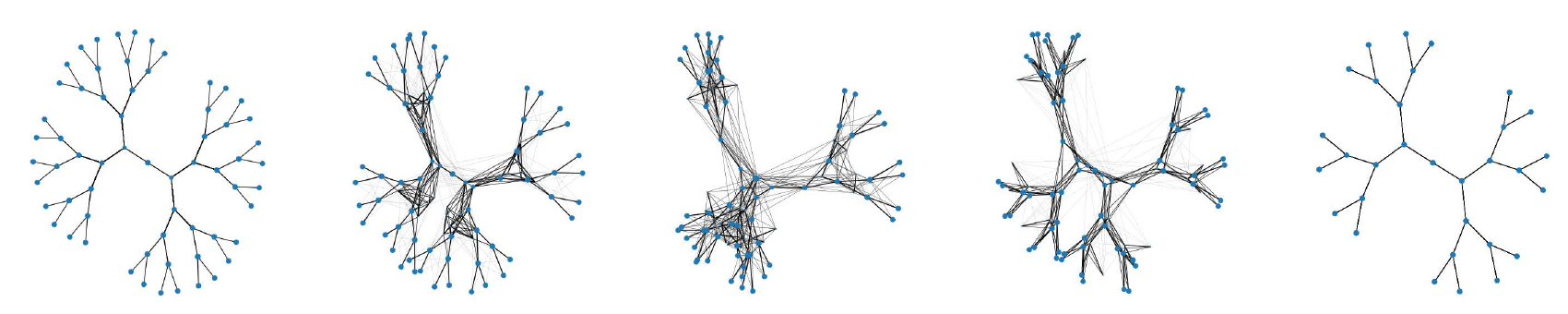}

\includegraphics[width = 0.45\textwidth]{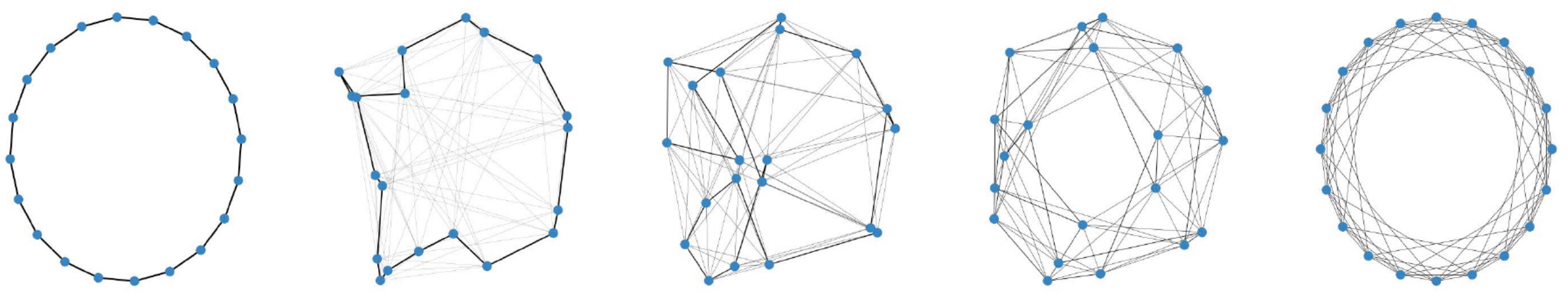}
\hspace{.15in}
\includegraphics[width = 0.45\textwidth]{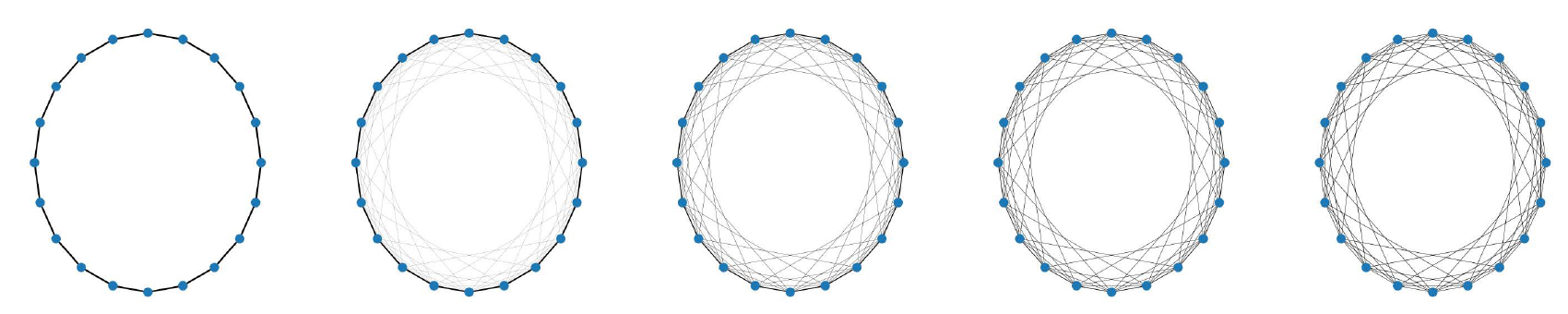}

\includegraphics[width = 0.45\textwidth]{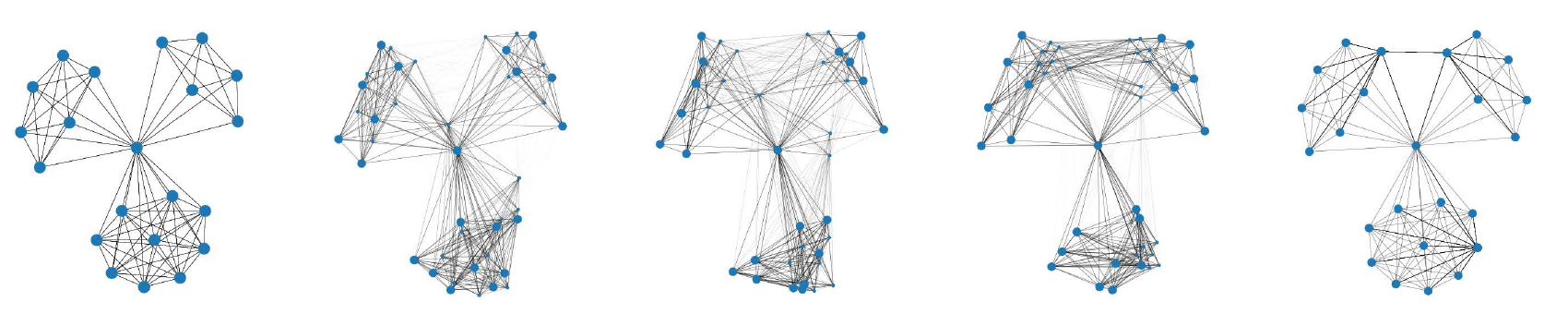}
\hspace{.15in}
\includegraphics[width = 0.45\textwidth]{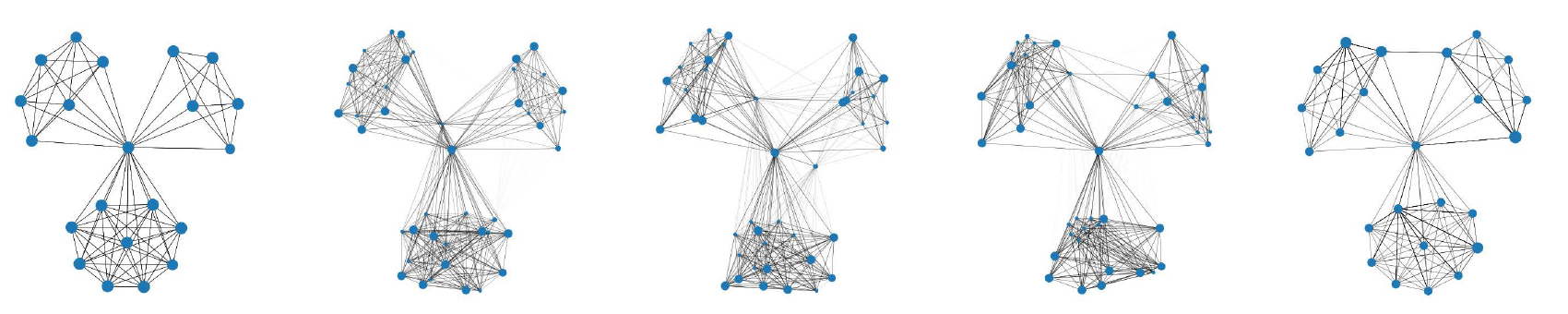}

\caption{Visualizations of GW graph matchings via adjacency loss \eqref{eqn:gwl_loss} (\textbf{left column}) and spectral loss \eqref{eqn:gwspec_loss} with $t = 20$ (\textbf{right column}). For each loss, we illustrate matchings between binary trees (\textbf{Top}), circular graphs (\textbf{Middle}) and IMDB graphs (\textbf{Bottom}). In each case, the interpolation is generated from node matchings inferred from the optimal GW coupling. Uniform node distributions are used in all examples.}\label{fig:matchings}
\end{figure*}

\paragraph{Graph Matching and Averaging.}\label{subsec:graph_matching}

Let $(G,p)$ and $(H,q)$ be measure graphs. Minimizers of the loss functions \eqref{eqn:gwl_loss} and \eqref{eqn:gwspec_loss} are couplings $C$ giving soft node correspondences between $G$ and $H$. To assess the intuitive meaning of such a coupling, it is useful to visualize this node correspondence at the graph level. We produce such visualizations using ideas from \cite{gwa}, as shown in Figure \ref{fig:matchings}. The figure shows six separate examples. For any particular example, we display an interpolation between graph $G$ (on the left) and graph $H$ (on the right). The optimal coupling $C$ is used to interpolate node positions from $G$ to $H$, with new edges phasing in during the interpolation---see Supplementary Materials for the details of the visualization algorithm as well as another experiment illustrating improved stability of the graph averaging algorithm of \cite{pcs16} when using spectral loss. We observe that in all cases, the matching produced via spectral loss preserves large scale qualitative features of the graphs more faithfully than adjacency-based matchings.

\begin{table}
    \centering
    \caption{Node Correctness, Mean $\pm$ St. Dev. (Time).}
    \label{tab:matching_scores}
    \begin{tabular}{ccc}
    \toprule
        Dataset  & $\mathrm{Adj}$ & $\mathrm{Spec}^{10}$ \\
    \midrule
       Proteins & .68 $\pm$ .22 (31.9) & \textbf{.78} $\pm$ .22 (5.1) \\
       Enzymes & .70 $\pm$ .18 (8.9) & \textbf{.79} $\pm$ .17 (1.4) \\
       Reddit & .29 $\pm$ .21 (3941.7) & \textbf{.50} $\pm$ .11 (206.1) \\
       Collab & \textbf{.50} $\pm$ .27 (4.3) & \textbf{.50} $\pm$ .27 (5.6)  \\
    \bottomrule
    \end{tabular}

\end{table}

We also assess the quality of node correspondences quantitatively. In this experiment, we consider two biological graph databases \textbf{Proteins} \citep{borgwardt2005protein} (1113 graphs with 39.06 nodes and 72.82 edges on average) and \textbf{Enzymes} \citep{dobson2003distinguishing, schomburg2004brenda} (600 graphs with 32.63 nodes and 62.14 edges on average), and two social graph databases \textbf{Reddit} (subset of 500 graphs with 375.9 nodes and 449.3 edges on average) and \textbf{Collab} (subset of 1000 graphs with 63.5 nodes and 855.6 edges on average), both from \cite{yanardag2015deep}. All datasets were downloaded from \cite{KKMMN2016}. For each graph $G = (V,E)$, we assign a node distribution $p$ from \eqref{eqn:distributions} (parameters tuned for the best performance in each method). A ``new'' measure graph $(H,q)$ is created by randomly permuting node labels of $G$. There is a ground truth node correspondence between $G$ and $H$ and the goal is to measure the ability of $\mathrm{GWL}$ and \gwspec to recover it. Given a coupling $C$ of $(G,p)$ and $(H,q)$, we measure its performance by the \emph{node correctness score} $|S \cap S_{GT}|/|S|$, where $S = \{(i,j) \mid C_{ij} > \epsilon \}$, for $\epsilon > 0$ a small threshold parameter, is the set of node correspondences from $C$ and $S_{GT}$ comprises ground truth node correspondences. \cite{xu2019scalable} showed that the adjacency-based GWL framework achieves state-of-the-art graph matching performance with respect to this metric.

For each graph $G$ and permuted version $H$, we compute couplings minimizing adjacency loss \eqref{eqn:gwl_loss} and spectral loss  \eqref{eqn:gwspec_loss} with $t = 10$ and compute their node correctness scores. Mean scores for each dataset are provided in Table \ref{tab:matching_scores}, where we see that spectral loss outperforms adjacency, except on the dense \textbf{Collab} graph where results agree. Minimizers of spectral loss in \gwspec were computed via standard gradient descent. Minimizers of adjacency loss in GWL were computed via both gradient descent and the regularized proximal gradient descent method of \cite{xu2019scalable}, with the best results we obtained reported in Table \ref{tab:matching_scores} (proximal for the biological graphs, standard for the social graphs). We have empirically found that $t = 10$ gives good matching performance for graphs with tens or hundreds of nodes, and moreover that performance is generally quite robust to this parameter choice (cf. illustration in Supplementary Materials).

\begin{figure*}
    \centering
    \includegraphics[width=\linewidth]{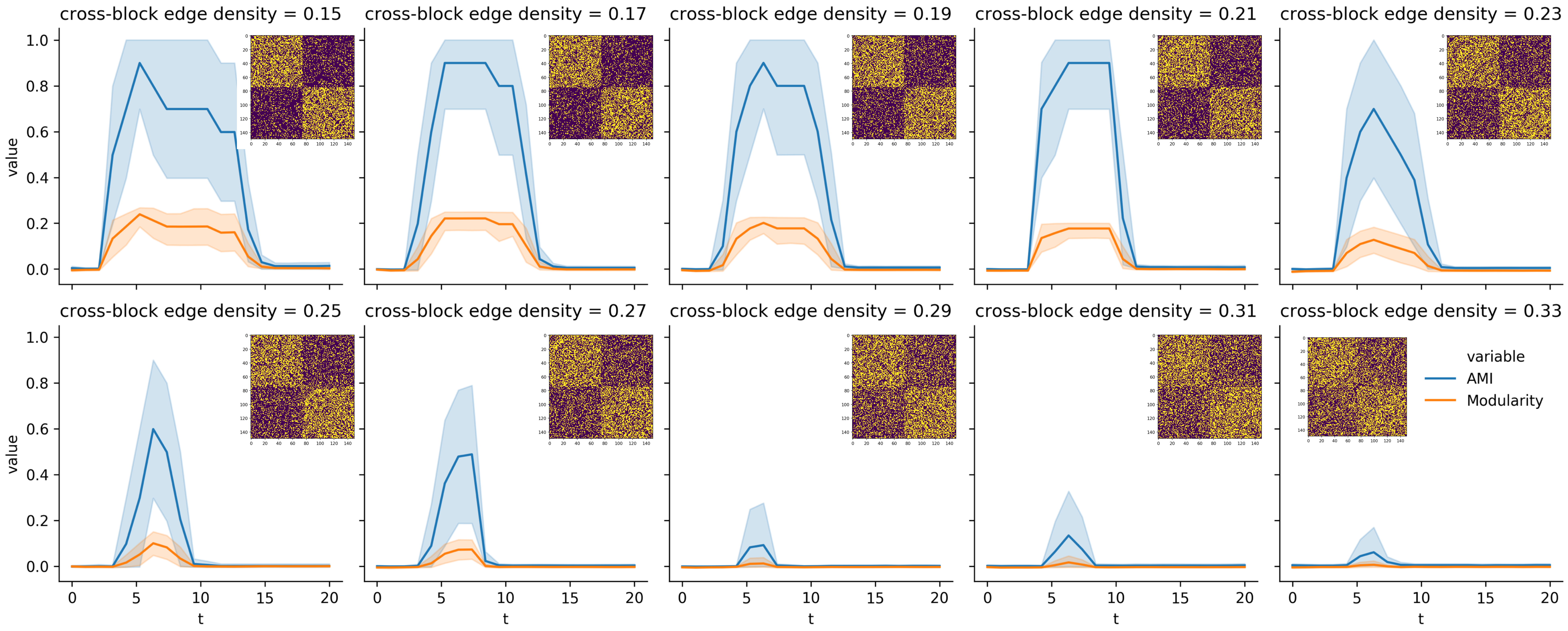}
    \caption{Plots of AMI and modularity for stochastic block model graphs across $t$ parameters with clearly correlated peaks. \textbf{Inset:} Sample adjacency matrices for SBMs with increasing cross-block densities.}
    \label{fig:partition_sbm_two}
\end{figure*}

\begin{table}[!t]
\caption{GRP model results; averaged AMI (time).}
\label{tab:GRP_results}
\begin{center}
\begin{tabular}{l c c c}
\toprule
$p_{out}$ & Infomap & GWL & \gwspec  \\\midrule
$.08$ & .676 (.34) & .611 (1.44) & \textbf{.845} (.62) \\
$.10$ & --- & .624 (1.56) & \textbf{.820} (.70) \\
$.12$ & --- &  .615 (1.64) &  \textbf{.820} (.75) \\
$.15$ & --- &  .611 (1.82) &  \textbf{.788} (.87) \\
\bottomrule
\end{tabular}
\end{center}
\end{table}

\paragraph{Optimizing the Scale Parameter.}
SpecGWL requires determining an appropriate scale parameter $t$, and here we propose two methods to this end in the context of graph partitioning.
We first propose a supervised method for tuning $t$ via \textbf{cross-validation}. We generate a dataset (illustration in Supplementary Materials) of $n=10$ stochastic block model networks with blocks of uniformly random sizes in the range $[20,50]$, within-block edge density $0.5$, and uniformly random across-block edge densities in the range $[0,0.35]$. Following a leave-one-out cross-validation scheme, we take $n$ subsets of the data and run $n$ trials. In trial $j$ we use subset $j$ as a training set to optimize the scale parameter $t$ via a grid search. Specifically, the optimal $t$ was the one that maximized the sum of the \emph{adjusted mutual information} (AMI) scores computed between the ground truth clusterings and the clusterings recovered by SpecGWL across the training set. We then evaluate the performance of SpecGWL on the test set according to the AMI score. Additionally we evaluate the following baselines: \textbf{Fluid} \citep{pares2017fluid}, \textbf{FastGreedy} \citep{clauset2004finding}, \textbf{Louvain} \citep{blondel2008fast}, \textbf{Infomap} \citep{rosvall2008maps}, and the adjacency-based GWL \citep{xu2019scalable}. Average AMI scores were: (Fluid, 0.59), (FastGreedy, 0.48), (Louvain, {\bf0.65}), (Infomap, 0.0), (GWL, 0.62), (SpecGWL, \textbf{0.65}), where the results for Fluid and Louvain were tuned via gridsearch over their parameters (number of communities and \emph{resolution}, respectively). Additionally, SpecGWL outperformed all other baselines in 5 out of 10 trials, with Louvain winning 4 and GWL winning the remaining trial. Moreover, SpecGWL outperformed GWL in 7 out of 10 trials.

We also propose a fully unsupervised method to tune the scale parameter via \textbf{modularity maximization}: here we use the \emph{Newman modularity Q-score} as a proxy for AMI---Figure \ref{fig:partition_sbm_two} illustrates how the peaks of AMI and modularity occur together across the scale parameter axis using synthetic data generated from a stochastic block model. In our unsupervised tuning method, the number of clusters $k$ is selected by maximizing the modularity (which requires no ground truth partition) of a \gwspec partitioning with respect to a fixed $t$-value (we use $t=10$ at this stage, which we  generally found to be stable to parameter perturbation for graphs on the order of 100s to 1000s of nodes) over a range of $k$-values. Once a $k$ has been selected, the scale parameter $t$ is also selected by maximizing modularity over a range of $t$ values. To illustrate this method with an unsupervised task, we test partitioning performance on Gaussian Random Partition model networks. This is a parametric model for random graphs, where the parameters are number of nodes (we use $1000$), mean number of nodes per cluster (we use $150$), a probability of within cluster edges (we use $0.5$) and a probability of out-of-cluster edges $p_{out} \in \{.08,.10,.12,.15\}$. For each value of $p_{out}$, we construct a random graph on which \gwspec is trained to determine number of clusters and optimal $t$-value. We then create 10 random graphs, partition them and compute AMI against ground truth clusters. Since this is a directed model, the only algorithms that apply are Infomap, GWL and \gwspec. Average AMI scores and compute times for each $p_{out}$ are reported in Table \ref{tab:GRP_results}.

\paragraph{Graph Partitioning on Real Data.}\label{subsec:graph_partitioning}

\begin{table*}
    \centering
    \caption{Comparison of adjusted mutual information scores across a variety of datasets.}
    \label{tab:ami}
    \begin{tabular}{
    p{0.1\textwidth} p{0.13\textwidth} 
    p{0.05\textwidth} p{0.1\textwidth}
    p{0.08\textwidth} p{0.08\textwidth}
    p{0.08\textwidth} p{0.1\textwidth}}
    \toprule 
    \multicolumn{2}{c}{Dataset} &
    Fluid & FastGreedy & Louvain & Infomap
    & GWL & {\gwspec}\\
    \midrule
    \multirow[t]{4}{*}{Wikipedia} & sym, raw & --- & 0.382 & 0.377 & 0.332 & 0.312 & $\textbf{0.442}^*$\\
    {} & sym, noisy & ---
    & 0.341 & 0.329 & 0.329 & 0.285 & \textbf{0.395}\\
    {} & asym, raw & --- & --- & ---
    & 0.332 & 0.178 & \textbf{0.376}\\
    {} & asym, noisy & --- & --- & --- & \textbf{0.329} & 0.170 & 0.307\\
    \midrule
    \multirow[t]{4}{*}{EU-email} & sym, raw & --- & 0.312 & 0.447 & 0.374 & 0.451 & \textbf{0.487}\\
    {} & sym, noisy & ---
    & 0.251 & 0.382 & 0.379 & 0.404 & \textbf{0.425}\\
    {} & asym, raw & --- & --- & ---
    & \textbf{0.443} & 0.420 & 0.437\\
    {} & asym, noisy & --- & --- & --- & 0.356 & \textbf{0.422} & 0.377\\
    \midrule
    \multirow[t]{2}{*}{Amazon} & raw & --- & 0.637 & 0.622 & \textbf{0.940} & $0.443^*$ & 0.692\\
    {} & noisy & 0.347
    & 0.573 & \textbf{0.584} & 0.463 & 0.352 & 0.441\\
    \midrule
    \multirow[t]{2}{*}{Village} & raw & --- & \textbf{0.881} & \textbf{0.881} & \textbf{0.881} & $0.606^*$ & $0.801^*$\\
    {} & noisy & --- & 0.778
    & \textbf{0.827} & 0.190 & 0.560 & 0.758 \\
    \bottomrule
    \end{tabular}
    
    {\small *Slight improvements possible with proximal gradient, but  overall performance rankings are preserved.}
\end{table*}

We compare the performance of \gwspec{} on a graph partitioning task against the baseline methods described earlier for four real-world datasets. The first is a directed \textbf{Wikipedia} hyperlink network \citep{snapnets} that we preprocessed by choosing 15 webpage categories and extracting their induced subgraphs. The resulting digraph had 1998 nodes and 2700 edges. The second was obtained from an \textbf{Amazon} product network \citep{snapnets} by taking the subgraph induced by the top 12 product categories. The resulting graph had 1501 nodes and 4626 edges. The third dataset was a digraph of email interactions between 42 departments of a European research institute (\textbf{EU-email}), and it comprised 1005 nodes and 25571 edges. The final dataset was a real-world network of interactions (8423 edges) among 1991 residents of 12 Indian villages \citep{banerjee2013diffusion}, which we refer to as the \textbf{Village} dataset. We created noisy versions of each graph by adding up to 10\% additional edges, and also created symmetrized versions of the Wikipedia and EU-email graphs by adding reciprocal edges.

The quality of each graph partition was measured by computing the AMI score against the ground-truth partition. The scores reported in Table \ref{tab:ami} are obtained using parameters selected through the unsupervised procedure described in the previous section. Reported results use standard gradient descent to compute GWL and SpecGWL scores. Despite issues with numerical instability, we also computed scores via the regularized proximal gradient method of \cite{xu2019scalable} where possible. Slight score improvements are possible in some cases, but overall score rankings between methods are unchanged---more experimental details are provided in the Supplementary Materials. We find that SpecGWL is the most consistent leader across all methods; in particular, it consistently produces improved results compared to GWL, the only other comprehensive method which is also applicable to graph matching and averaging. The performance of SpecGWL on directed graphs is especially relevant, considering that its closest competitor Infomap is a state-of-the-art method for digraph partitioning.

\section{DISCUSSION}

We have introduced a spectral notion of GW distance for graph comparison problems based on comparing heat kernels rather than adjacency matrices. This spectral variant is shown qualitatively and quantitatively to improve performance in graph matching and partitioning tasks. The techniques introduced here should be useful for studying further variants of GW distances. For example, work of \cite{pcs16} suggests that replacing the $L^2$-type spectral loss \eqref{eqn:gwspec_loss} with a loss based on KL divergence could have benefits when performing statistical analysis on graph datasets. We also remark that, while spectral GW is faster than its adjacency counterpart on smaller graphs, it may not enjoy the same scalability properties due to the lack of sparse matrix operations. 
A significant direction of future work will be to construct multiscale approaches to analyze large scale graphs through both adjacency and spectral methods: the divide-and-conquer techniques of \cite{xu2019scalable} can break graphs into manageable chunks, and then SpecGWL can finish the computation with improved runtime and accuracy.
On the theoretical front, Theorems \ref{thm:sparse-coup} and \ref{thm:fiedler_partition} suggest that it is tractable to study spectral GW rigorously, and developing Theorem \ref{thm:fiedler_partition} into a larger theory should illuminate further connections between optimal transport and spectral graph theory. It will also be interesting to explore more fully the dependence of our algorithms on the scale parameter $t$ and to incorporate multiple $t$-values into computations.

\paragraph{Acknowledgments} We would like to thank the reviewers on this work for their numerous helpful comments. We wish to also thank Facundo Mémoli for suggesting references related to the original work on Spectral GW.

\bibliographystyle{plainnat}
\bibliography{biblio}

\begin{thebibliography}{54}
\providecommand{\natexlab}[1]{#1}
\providecommand{\url}[1]{\texttt{#1}}
\expandafter\ifx\csname urlstyle\endcsname\relax
  \providecommand{\doi}[1]{doi: #1}\else
  \providecommand{\doi}{doi: \begingroup \urlstyle{rm}\Url}\fi

\bibitem[Alvarez-Melis and Jaakkola(2018)]{alvarez2018gromov}
David Alvarez-Melis and Tommi Jaakkola.
\newblock {G}romov-{W}asserstein alignment of word embedding spaces.
\newblock In \emph{Proceedings of the 2018 Conference on Empirical Methods in
  Natural Language Processing}, pages 1881--1890, 2018.

\bibitem[Alvarez-Melis et~al.(2019)Alvarez-Melis, Jegelka, and
  Jaakkola]{alvarez2019towards}
David Alvarez-Melis, Stefanie Jegelka, and Tommi~S Jaakkola.
\newblock Towards optimal transport with global invariances.
\newblock In \emph{The 22nd International Conference on Artificial Intelligence
  and Statistics}, pages 1870--1879. PMLR, 2019.

\bibitem[Banerjee et~al.(2013)Banerjee, Chandrasekhar, Duflo, and
  Jackson]{banerjee2013diffusion}
Abhijit Banerjee, Arun~G Chandrasekhar, Esther Duflo, and Matthew~O Jackson.
\newblock The diffusion of microfinance.
\newblock \emph{Science}, 341\penalty0 (6144):\penalty0 1236498, 2013.

\bibitem[Barbe et~al.(2020)Barbe, Sebban, Gon{\c{c}}alves, Borgnat, and
  Gribonval]{barbe2020graph}
Am{\'e}lie Barbe, Marc Sebban, Paulo Gon{\c{c}}alves, Pierre Borgnat, and
  R{\'e}mi Gribonval.
\newblock Graph diffusion {W}asserstein distances.
\newblock In \emph{European Conference on Machine Learning and Principles and
  Practice of Knowledge Discovery in Databases}, 2020.

\bibitem[Blondel et~al.(2008)Blondel, Guillaume, Lambiotte, and
  Lefebvre]{blondel2008fast}
Vincent~D Blondel, Jean-Loup Guillaume, Renaud Lambiotte, and Etienne Lefebvre.
\newblock Fast unfolding of communities in large networks.
\newblock \emph{Journal of statistical mechanics: theory and experiment},
  2008\penalty0 (10):\penalty0 P10008, 2008.

\bibitem[Borgwardt et~al.(2005)Borgwardt, Ong, Sch{\"o}nauer, Vishwanathan,
  Smola, and Kriegel]{borgwardt2005protein}
Karsten~M Borgwardt, Cheng~Soon Ong, Stefan Sch{\"o}nauer, SVN Vishwanathan,
  Alex~J Smola, and Hans-Peter Kriegel.
\newblock Protein function prediction via graph kernels.
\newblock \emph{Bioinformatics}, 21\penalty0 (suppl\_1):\penalty0 i47--i56,
  2005.

\bibitem[Bronstein and Glashoff(2013)]{bronstein2013heat}
Michael~M Bronstein and Klaus Glashoff.
\newblock Heat kernel coupling for multiple graph analysis.
\newblock \emph{arXiv preprint arXiv:1312.3035}, 2013.

\bibitem[Bunne et~al.(2019)Bunne, Alvarez-Melis, Krause, and
  Jegelka]{bunne2019learning}
Charlotte Bunne, David Alvarez-Melis, Andreas Krause, and Stefanie Jegelka.
\newblock Learning generative models across incomparable spaces.
\newblock In \emph{International Conference on Machine Learning}, pages
  851--861, 2019.

\bibitem[Chapel et~al.(2020)Chapel, Alaya, and Gasso]{chapel2020partial}
Laetitia Chapel, Mokhtar~Z Alaya, and Gilles Gasso.
\newblock Partial optimal tranport with applications on positive-unlabeled
  learning.
\newblock \emph{Advances in Neural Information Processing Systems}, 33, 2020.

\bibitem[Chizat(2017)]{chizat-thesis}
L{\'e}na{\"\i}c Chizat.
\newblock \emph{Transport optimal de mesures positives: mod{\`e}les,
  m{\'e}thodes num{\'e}riques, applications}.
\newblock PhD thesis, Universit\'{e} Paris-Dauphine, 2017.

\bibitem[Chowdhury and M{\'e}moli(2019)]{gwnets}
Samir Chowdhury and Facundo M{\'e}moli.
\newblock The {G}romov--{W}asserstein distance between networks and stable
  network invariants.
\newblock \emph{Information and Inference: A Journal of the IMA}, 8\penalty0
  (4):\penalty0 757--787, 2019.

\bibitem[Chowdhury and Needham(2020)]{gwa}
Samir Chowdhury and Tom Needham.
\newblock Gromov-{W}asserstein averaging in a {R}iemannian framework.
\newblock In \emph{Proceedings of the IEEE/CVF Conference on Computer Vision
  and Pattern Recognition Workshops}, pages 842--843, 2020.

\bibitem[Chung(2005)]{chung2005laplacians}
Fan Chung.
\newblock Laplacians and the {C}heeger inequality for directed graphs.
\newblock \emph{Annals of Combinatorics}, 9\penalty0 (1):\penalty0 1--19, 2005.

\bibitem[Clauset et~al.(2004)Clauset, Newman, and Moore]{clauset2004finding}
Aaron Clauset, Mark~EJ Newman, and Cristopher Moore.
\newblock Finding community structure in very large networks.
\newblock \emph{Physical review E}, 70\penalty0 (6):\penalty0 066111, 2004.

\bibitem[De~Ponti and Mondino(2020)]{de2020entropy}
Nicol{\'o} De~Ponti and Andrea Mondino.
\newblock Entropy-transport distances between unbalanced metric measure spaces.
\newblock \emph{arXiv preprint arXiv:2009.10636}, 2020.

\bibitem[Dobson and Doig(2003)]{dobson2003distinguishing}
Paul~D Dobson and Andrew~J Doig.
\newblock Distinguishing enzyme structures from non-enzymes without alignments.
\newblock \emph{Journal of molecular biology}, 330\penalty0 (4):\penalty0
  771--783, 2003.

\bibitem[Dong and Sawin(2020)]{dong2020copt}
Yihe Dong and Will Sawin.
\newblock {COPT}: Coordinated optimal transport on graphs.
\newblock \emph{Advances in Neural Information Processing Systems}, 33, 2020.

\bibitem[Fiedler(1973)]{fiedler1973algebraic}
Miroslav Fiedler.
\newblock Algebraic connectivity of graphs.
\newblock \emph{Czechoslovak mathematical journal}, 23\penalty0 (2):\penalty0
  298--305, 1973.

\bibitem[Flamary and Courty(2017)]{pot}
R\'{e}mi Flamary and Nicolas Courty.
\newblock {POT}: {P}ython {O}ptimal {T}ransport library.
\newblock 2017.
\newblock URL \url{https://github.com/rflamary/POT}.

\bibitem[Girvan and Newman(2002)]{girvan2002community}
Michelle Girvan and Mark~EJ Newman.
\newblock Community structure in social and biological networks.
\newblock \emph{Proceedings of the National Academy of Sciences}, 99\penalty0
  (12):\penalty0 7821--7826, 2002.

\bibitem[Hendrikson(2016)]{hendrikson}
Reigo Hendrikson.
\newblock Using {G}romov-{W}asserstein distance to explore sets of networks.
\newblock Master's thesis, University of Tartu, 2016.

\bibitem[Hu et~al.(2014)Hu, Rustamov, and Guibas]{hu2014stable}
Nan Hu, Raif~M Rustamov, and Leonidas Guibas.
\newblock Stable and informative spectral signatures for graph matching.
\newblock In \emph{Proceedings of the IEEE Conference on Computer Vision and
  Pattern Recognition}, pages 2305--2312, 2014.

\bibitem[Kasue and Kumura(1994)]{kasue1994spectral}
Atsushi Kasue and Hironori Kumura.
\newblock Spectral convergence of {R}iemannian manifolds.
\newblock \emph{Tohoku Mathematical Journal, Second Series}, 46\penalty0
  (2):\penalty0 147--179, 1994.

\bibitem[Kersting et~al.(2016)Kersting, Kriege, Morris, Mutzel, and
  Neumann]{KKMMN2016}
Kristian Kersting, Nils~M. Kriege, Christopher Morris, Petra Mutzel, and Marion
  Neumann.
\newblock Benchmark data sets for graph kernels, 2016.
\newblock URL \url{http://graphkernels.cs.tu-dortmund.de}.

\bibitem[Kolouri et~al.(2017)Kolouri, Park, Thorpe, Slepcev, and
  Rohde]{kolouri2017optimal}
Soheil Kolouri, Se~Rim Park, Matthew Thorpe, Dejan Slepcev, and Gustavo~K
  Rohde.
\newblock Optimal mass transport: Signal processing and machine-learning
  applications.
\newblock \emph{IEEE signal processing magazine}, 34\penalty0 (4):\penalty0
  43--59, 2017.

\bibitem[Leskovec and Krevl(2014)]{snapnets}
Jure Leskovec and Andrej Krevl.
\newblock {SNAP Datasets}: {Stanford} large network dataset collection.
\newblock \url{http://snap.stanford.edu/data}, June 2014.

\bibitem[M{\'e}moli(2007)]{dgh-sm}
Facundo M{\'e}moli.
\newblock On the use of {G}romov-{H}ausdorff distances for shape comparison.
\newblock \emph{The Eurographics Association}, 2007.

\bibitem[M{\'e}moli(2011{\natexlab{a}})]{dghlp-focm}
Facundo M{\'e}moli.
\newblock Gromov--{W}asserstein distances and the metric approach to object
  matching.
\newblock \emph{Foundations of Computational Mathematics}, 11\penalty0
  (4):\penalty0 417--487, 2011{\natexlab{a}}.

\bibitem[M{\'e}moli(2011{\natexlab{b}})]{memoli2011spectral}
Facundo M{\'e}moli.
\newblock A spectral notion of {G}romov--{W}asserstein distance and related
  methods.
\newblock \emph{Applied and Computational Harmonic Analysis}, 30\penalty0
  (3):\penalty0 363--401, 2011{\natexlab{b}}.

\bibitem[M{\'e}moli and Needham(2018)]{memoli2018gromov}
Facundo M{\'e}moli and Tom Needham.
\newblock {G}romov-{M}onge quasi-metrics and distance distributions.
\newblock \emph{arXiv preprint arXiv:1810.09646}, 2018.

\bibitem[Nassar et~al.(2018)Nassar, Veldt, Mohammadi, Grama, and
  Gleich]{nassar2018low}
Huda Nassar, Nate Veldt, Shahin Mohammadi, Ananth Grama, and David~F Gleich.
\newblock Low rank spectral network alignment.
\newblock In \emph{Proceedings of the 2018 World Wide Web Conference}, pages
  619--628, 2018.

\bibitem[Par{\'e}s et~al.(2017)Par{\'e}s, Gasulla, Vilalta, Moreno,
  Ayguad{\'e}, Labarta, Cort{\'e}s, and Suzumura]{pares2017fluid}
Ferran Par{\'e}s, Dario~Garcia Gasulla, Armand Vilalta, Jonatan Moreno, Eduard
  Ayguad{\'e}, Jes{\'u}s Labarta, Ulises Cort{\'e}s, and Toyotaro Suzumura.
\newblock Fluid communities: a competitive, scalable and diverse community
  detection algorithm.
\newblock In \emph{International Conference on Complex Networks and their
  Applications}, pages 229--240. Springer, 2017.

\bibitem[Patro and Kingsford(2012)]{patro2012global}
Rob Patro and Carl Kingsford.
\newblock Global network alignment using multiscale spectral signatures.
\newblock \emph{Bioinformatics}, 28\penalty0 (23):\penalty0 3105--3114, 2012.

\bibitem[Peyr{\'e} et~al.(2016)Peyr{\'e}, Cuturi, and Solomon]{pcs16}
Gabriel Peyr{\'e}, Marco Cuturi, and Justin Solomon.
\newblock Gromov-{W}asserstein averaging of kernel and distance matrices.
\newblock In \emph{International Conference on Machine Learning}, pages
  2664--2672, 2016.

\bibitem[Reuter et~al.(2006)Reuter, Wolter, and Peinecke]{reuter2006laplace}
Martin Reuter, Franz-Erich Wolter, and Niklas Peinecke.
\newblock Laplace--{B}eltrami spectra as ‘{S}hape-{DNA}’of surfaces and
  solids.
\newblock \emph{Computer-Aided Design}, 38\penalty0 (4):\penalty0 342--366,
  2006.

\bibitem[Rosvall and Bergstrom(2008)]{rosvall2008maps}
Martin Rosvall and Carl~T Bergstrom.
\newblock Maps of random walks on complex networks reveal community structure.
\newblock \emph{Proceedings of the National Academy of Sciences}, 105\penalty0
  (4):\penalty0 1118--1123, 2008.

\bibitem[Schmitzer and Schn{\"o}rr(2013)]{schmitzer2013modelling}
Bernhard Schmitzer and Christoph Schn{\"o}rr.
\newblock Modelling convex shape priors and matching based on the
  {G}romov-{W}asserstein distance.
\newblock \emph{Journal of mathematical imaging and vision}, 46\penalty0
  (1):\penalty0 143--159, 2013.

\bibitem[Schomburg et~al.(2004)Schomburg, Chang, Ebeling, Gremse, Heldt, Huhn,
  and Schomburg]{schomburg2004brenda}
Ida Schomburg, Antje Chang, Christian Ebeling, Marion Gremse, Christian Heldt,
  Gregor Huhn, and Dietmar Schomburg.
\newblock Brenda, the enzyme database: updates and major new developments.
\newblock \emph{Nucleic acids research}, 32\penalty0 (suppl\_1):\penalty0
  D431--D433, 2004.

\bibitem[S{\'e}journ{\'e} et~al.(2020)S{\'e}journ{\'e}, Vialard, and
  Peyr{\'e}]{sejourne2020unbalanced}
Thibault S{\'e}journ{\'e}, Fran{\c{c}}ois-Xavier Vialard, and Gabriel
  Peyr{\'e}.
\newblock The {U}nbalanced {G}romov {W}asserstein distance: Conic formulation
  and relaxation.
\newblock \emph{arXiv preprint arXiv:2009.04266}, 2020.

\bibitem[Smith(1984)]{smith1984efficient}
Robert~L Smith.
\newblock Efficient {M}onte {C}arlo procedures for generating points uniformly
  distributed over bounded regions.
\newblock \emph{Operations Research}, 32\penalty0 (6):\penalty0 1296--1308,
  1984.

\bibitem[Solomon et~al.(2016)Solomon, Peyr{\'e}, Kim, and Sra]{s16}
Justin Solomon, Gabriel Peyr{\'e}, Vladimir~G Kim, and Suvrit Sra.
\newblock Entropic metric alignment for correspondence problems.
\newblock \emph{ACM Transactions on Graphics (TOG)}, 35\penalty0 (4):\penalty0
  72, 2016.

\bibitem[Sturm(2006)]{sturm2006geometry}
Karl-Theodor Sturm.
\newblock On the geometry of metric measure spaces.
\newblock \emph{Acta mathematica}, 196\penalty0 (1):\penalty0 65--131, 2006.

\bibitem[Sturm(2012)]{sturm2012space}
Karl-Theodor Sturm.
\newblock The space of spaces: curvature bounds and gradient flows on the space
  of metric measure spaces.
\newblock \emph{arXiv preprint arXiv:1208.0434}, 2012.

\bibitem[Sun et~al.(2009)Sun, Ovsjanikov, and Guibas]{sun2009concise}
Jian Sun, Maks Ovsjanikov, and Leonidas Guibas.
\newblock A concise and provably informative multi-scale signature based on
  heat diffusion.
\newblock In \emph{Computer graphics forum}, volume~28, pages 1383--1392. Wiley
  Online Library, 2009.

\bibitem[Tsitsulin et~al.(2018)Tsitsulin, Mottin, Karras, Bronstein, and
  M{\"u}ller]{netlsd}
Anton Tsitsulin, Davide Mottin, Panagiotis Karras, Alexander Bronstein, and
  Emmanuel M{\"u}ller.
\newblock {NetLSD}: hearing the shape of a graph.
\newblock In \emph{Proceedings of the 24th ACM SIGKDD International Conference
  on Knowledge Discovery \& Data Mining}, pages 2347--2356. ACM, 2018.

\bibitem[Tsitsulin et~al.(2020)Tsitsulin, Munkhoeva, Mottin, Karras, Bronstein,
  Oseledets, and M{\"u}ller]{tsitsulin2020shape}
Anton Tsitsulin, Marina Munkhoeva, Davide Mottin, Panagiotis Karras, Alex
  Bronstein, Ivan Oseledets, and Emmanuel M{\"u}ller.
\newblock The shape of data: Intrinsic distance for data distributions.
\newblock In \emph{ICLR 2020: Proceedings of the International Conference on
  Learning Representations}, 2020.

\bibitem[Vayer et~al.(2018)Vayer, Chapel, Flamary, Tavenard, and
  Courty]{vayer2018fused}
Titouan Vayer, Laetita Chapel, R{\'e}mi Flamary, Romain Tavenard, and Nicolas
  Courty.
\newblock Fused {G}romov-{W}asserstein distance for structured objects:
  theoretical foundations and mathematical properties.
\newblock \emph{arXiv preprint arXiv:1811.02834}, 2018.

\bibitem[Vayer et~al.(2019{\natexlab{a}})Vayer, Courty, Tavenard, and
  Flamary]{vayer2019optimal}
Titouan Vayer, Nicolas Courty, Romain Tavenard, and R{\'e}mi Flamary.
\newblock Optimal transport for structured data with application on graphs.
\newblock In \emph{International Conference on Machine Learning}, pages
  6275--6284, 2019{\natexlab{a}}.

\bibitem[Vayer et~al.(2019{\natexlab{b}})Vayer, Flamary, Courty, Tavenard, and
  Chapel]{vayer2019sliced}
Titouan Vayer, R{\'e}mi Flamary, Nicolas Courty, Romain Tavenard, and Laetitia
  Chapel.
\newblock Sliced {G}romov-{W}asserstein.
\newblock In \emph{Advances in Neural Information Processing Systems}, pages
  14726--14736, 2019{\natexlab{b}}.

\bibitem[Xu(2020)]{xu2020gromov}
Hongteng Xu.
\newblock Gromov-{W}asserstein factorization models for graph clustering.
\newblock In \emph{AAAI}, pages 6478--6485, 2020.

\bibitem[Xu et~al.(2019{\natexlab{a}})Xu, Luo, and Carin]{xu2019scalable}
Hongteng Xu, Dixin Luo, and Lawrence Carin.
\newblock Scalable {G}romov-{W}asserstein learning for graph partitioning and
  matching.
\newblock In \emph{Advances in Neural Information Processing Systems}, pages
  3046--3056, 2019{\natexlab{a}}.

\bibitem[Xu et~al.(2019{\natexlab{b}})Xu, Luo, Zha, and Carin]{xu2019gromov}
Hongteng Xu, Dixin Luo, Hongyuan Zha, and Lawrence Carin.
\newblock {G}romov-{W}asserstein learning for graph matching and node
  embedding.
\newblock In \emph{International Conference on Machine Learning}, pages
  6932--6941, 2019{\natexlab{b}}.

\bibitem[Xu et~al.(2020)Xu, Luo, Henao, Shah, and Carin]{xu2020learning}
Hongteng Xu, Dixin Luo, Ricardo Henao, Svati Shah, and Lawrence Carin.
\newblock Learning autoencoders with relational regularization.
\newblock In \emph{International Conference on Machine Learning}, pages
  10576--10586. PMLR, 2020.

\bibitem[Yanardag and Vishwanathan(2015)]{yanardag2015deep}
Pinar Yanardag and SVN Vishwanathan.
\newblock Deep graph kernels.
\newblock In \emph{Proceedings of the 21th ACM SIGKDD International Conference
  on Knowledge Discovery and Data Mining}, pages 1365--1374, 2015.

\end{thebibliography}

\newpage
\onecolumn
\appendix

\section{Proofs of Theorems}

\subsection{Theorem \ref{thm:sparse-coup}}

\begin{proof} Suppose $G=(V^G,E^G,p)$ and $H=(V^H,E^H,q)$ have $m,n$ nodes, respectively. Let $C \in \mathcal{C}(p,q)$. Then $C \in [0,1]^{m \times n}$ and satisfies $(m+n-1)$ linear equality constraints coming from the row and column sums. By Lemma \ref{lem:concave}, at least one of the minimizers of spectral loss \eqref{eqn:gwspec_loss} is located at an extreme point of the convex polytope $\mathcal{C}(p,q)$. This polytope lies in an $(mn - (m+n-1))$-dimensional affine subspace of $mn$-dimensional space. The equality constraints automatically ensure that each $C_{ij} < 1$, where the strict inequality holds because the graphs are fully supported and thus each $p_i, q_j < 1$. Therefore, estimating the number of zero entries is equivalent to estimating the number $k$ of active nonnegativity constraints. An extreme point corresponds to the intersection of $k$ hyperplanes in general position with this affine subspace, and this intersection has dimension $mn - (m+n-1) - k$. Because the extreme point has dimension 0, we have $k = mn - (m+n-1)$. If the hyperplanes are not in general position, then the number of active nonnegativity constraints, i.e. the number of zeros, is greater than or equal to $mn - (m+n-1)$. Next suppose $m \sim n$. Then the ratio of nonzero entries to total entries of $C$ is roughly $\tfrac{n^2 - k}{n^2} = \tfrac{2n-1}{n^2},$ and this term tends to 0 as $n \to \infty$. \qedhere
\end{proof}

\subsection{Theorem \ref{thm:fiedler_partition}}

\begin{proof}
Let $G=(V,E)$, with $|V|=n$, be a graph satisfying the assumptions, endowed with uniform vertex distribution $p$ and let $K^t$ denote the heat kernel of $G$. By definition,
\[
K^t = \Phi e^{-t \Lambda} \Phi^T = \sum_{j=1}^n e^{-t\lambda_j} \phi_j \phi_j^T,
\]
where $\Phi$ is a matrix whose columns are the orthonormal eigenvectors $\phi_1,\ldots,\phi_n$ of the graph Laplacian $L$ of $G$ and $\Lambda$ is the diagonal matrix of sorted eigenvalues $0 = \lambda_1 < \lambda_2 < \lambda_3 \leq \lambda_4 \leq \cdots \leq \lambda_n$ of $L$. Let $Q$ be the 2-way partitioning template from  \eqref{eqn:gwl_partition_loss}. Since $p$ is uniform, the estimated distribution $q$ is also uniform and $Q = \frac{1}{2} I_2$, where $I_2$ is the $2 \times 2$ identity matrix. 

The 2-way spectral GW partitioning of $G$ is obtained from a coupling minimizing the spectral partitioning loss \eqref{eqn:spec_gwl_partition_loss}. Using Lemma \ref{lem:concave}, we see that this optimization task is equivalent to maximizing
\[
C \mapsto \langle K^t C, C Q\rangle = \frac{1}{2}\langle K^t C, C \rangle
\]
over the coupling polytope $\mathcal{C}(p,q)$. Since the factor of $\frac{1}{2}$ does not effect the optimization, we supress it and further simplify the objective function as
\[
   \langle K^t C, C \rangle =  \mathrm{tr}\left((K^t C)^T C \right) = \mathrm{tr}\left(C C^T \sum_{j=1}^n e^{-t \lambda_j} \phi_j \phi_j^T \right) =  \sum_{j=1}^n e^{-t \lambda_j} \mathrm{tr}\left(CC^T \phi_j \phi_j^T\right).
\]
Since the leading eigenvector is $\phi_1 = \frac{1}{\sqrt{n}} 1^{n \times 1}$ (the normalized vector of all ones), it is easy to check that the term  $\mathrm{tr}(CC^T \phi_1 \phi_1^T)$ is constant for all $C \in \mathcal{C}(p,q)$. The objective therefore becomes to maximize over $C \in \mathcal{C}(p,q)$ the quantity
\[
\sum_{j=2}^n e^{-t \lambda_j} \mathrm{tr}\left(CC^T \phi_j \phi_j^T\right) = e^{-t \lambda_2} \left(\mathrm{tr}\left(CC^T \phi_2 \phi_2^T\right) + \sum_{j=3}^n e^{-t(\lambda_j - \lambda_2)} \mathrm{tr}\left(CC^T \phi_j \phi_j^T\right)\right),
\]
which is in turn equivalent to maximizing
\begin{equation}\label{eqn:partition_optimization}
C \mapsto \mathrm{tr}\left(CC^T \phi_2 \phi_2^T\right) + \sum_{j=3}^n e^{-t(\lambda_j - \lambda_2)} \mathrm{tr}\left(CC^T \phi_j \phi_j^T\right).
\end{equation}

Observe that the summation term goes to zero as $t \rightarrow \infty$ (using $\lambda_j > \lambda_2$ for all $j \geq 3$), while the first term is independent of $t$. It follows that, for sufficiently large $t$, maximization of \eqref{eqn:partition_optimization} is equivalent to maximizing
\begin{equation}\label{eqn:partition_optimization_2}
C \mapsto \mathrm{tr}\left(CC^T \phi_2 \phi_2^T\right)
\end{equation}
over $\mathcal{C}(p,q)$. It then remains to study the structure of maximizers of \eqref{eqn:partition_optimization_2}.

We further simplify the objective function \eqref{eqn:partition_optimization_2} as
\[
    \mathrm{tr}\left(CC^T \phi_2 \phi_2^T\right) = \mathrm{tr}\left((C^T\phi_2)^T (C^T \phi_2)\right) = \|C^T \phi_2\|^2,
\]
where the norm in the last line is the Frobenius norm. We denote the column vectors of $C$ by $C_1, C_2 \in \R^{1 \times n}$, so that 
\[
\|C^T \phi_2\|^2 = \left\|\left(\begin{array}{c}
C_1 \cdot \phi_2 \\
C_2 \cdot \phi_2 \end{array}\right) \right\|^2,
\]
where the norm on the right is the Euclidean norm. 
Since $C \in \mathcal{C}(p,q)$ and $p$ is uniform,  we have $C_2 = \frac{1}{n}1^{n \times 1} - C_1$, whence 
\[
C_2 \cdot \phi_2 = \left(\frac{1}{n}1^{n \times 1} - C_1\right) \cdot \phi_2 = -C_1 \cdot \phi_2,
\]
since $\phi_2$ is orthogonal to $1^{n \times 1} = \sqrt{n} \phi_1$. The objective \eqref{eqn:partition_optimization_2} is finally reduced to 
\begin{equation}\label{eqn:partition_optimization_3}
C \mapsto 2 (C_1 \cdot \phi_2)^2.
\end{equation}
Let $\phi_2^+$ be the vector of positive entries of $\phi_2$ with all negative entries thresholded to zero and likewise define $\phi_2^-$ to be the vector of negative entries of $\phi_2$. Assume without loss of generality that $\|\phi_2^+\| \geq \|\phi_2^-\|$ (the other case follows entirely similarly). Then in order to maximize \eqref{eqn:partition_optimization_3}, one should set each entry of $C_1$ to be nonzero if and only if the corresponding entry of $\phi_2$ is positive. The spectral GW partitioning therefore agrees with the Fiedler partitioning, and the proof is complete.
\end{proof}

\section{An MCMC Sampler for Couplings.}

Both the adjacency \eqref{eqn:gwl_loss} and spectral \eqref{eqn:gwspec_loss} loss functions are nonconvex, and solving such problems effectively often relies on a clever choice of initialization. A limitation of the current practice is that this initialization is often chosen to be the \emph{product coupling} $pq^T$, which we empirically find to be sub-optimal in even simple cases. This is accomplished by running gradient descent from each point in an ensemble of initializations generated by a Markov Chain Monte Carlo Hit-And-Run sampler \citep{smith1984efficient}. This algorithm is well-known, but we describe it below for the convenience of the reader. Our code includes a lean Python implementation written specifically for sampling the coupling polytope; we hope such an implementation will be useful to the broader optimal transport community.

\begin{center}
\begin{minipage}{0.9\textwidth}
\begin{algorithm}[H]
\caption{Markov chain sampler}\label{alg:markov}
\begin{algorithmic}[1]
\Function{markovStep}{$A,p,q,C$}
\State \emph{// $A$: matrix of linear constraints}
\State \emph{// $p,q: m\times 1, n \times 1$ probability vectors}
\State \emph{// $C: m\times n$ initial coupling matrix} \\
\State $V \gets$ random $m\times n$ matrix as direction
\State $Q \gets$ o.n. basis for row space of $A$
\State $V\gets V - Q Q^T V$ \Comment{project $V$ to correct subspace}
\State $\mathrm{pos} \gets$ indices where $V > 0$
\State $\mathrm{neg} \gets$ indices where $V < 0$
\State $\alpha \gets \mathrm{max}(-C[\mathrm{pos}]/V[\mathrm{pos}])$
\State $\beta \gets \mathrm{min}(-C[\mathrm{neg}]/V[\mathrm{neg}])$
\Comment{$[\alpha,\beta]$ is maximal range of step sizes}
\State $\gamma \gets$ random element of $[\alpha,\beta]$ \\
\Return $C + \gamma V$ \Comment{new coupling matrix}
\EndFunction
\end{algorithmic}
\end{algorithm}
\end{minipage}
\end{center}

\section{Additional experiments and implementation details}
\label{subsec:exp-details}

\subsection{Additional Landscape Results}

\begin{figure}
\centering
\includegraphics[width = 0.48\textwidth]{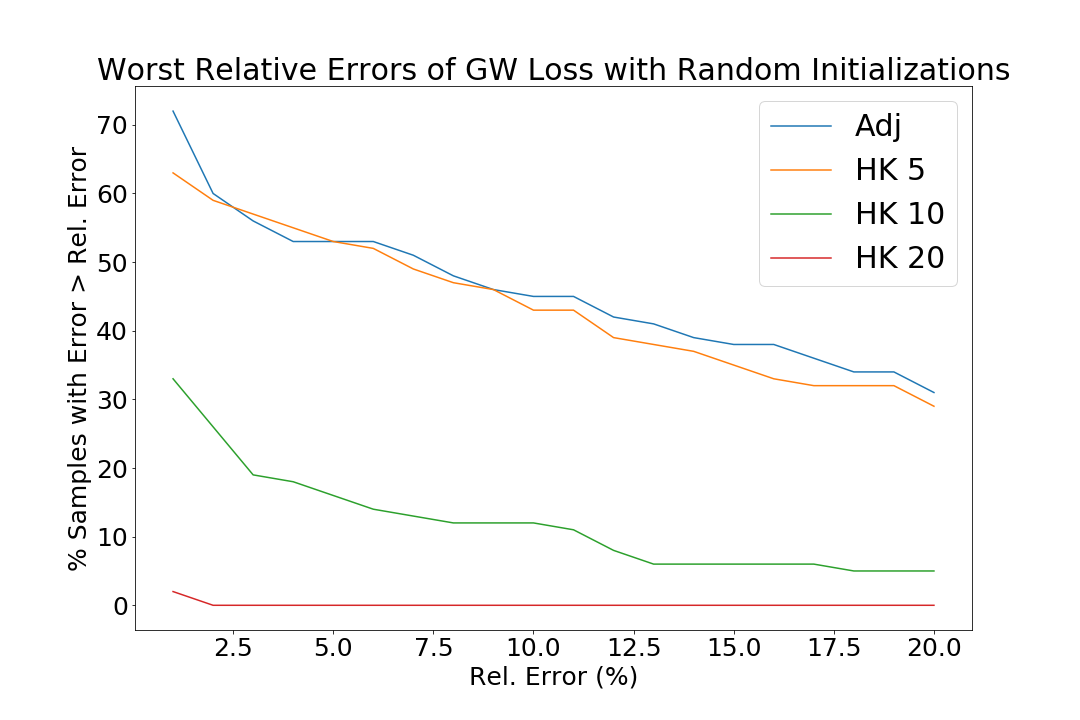}
\includegraphics[width = 0.48\textwidth]{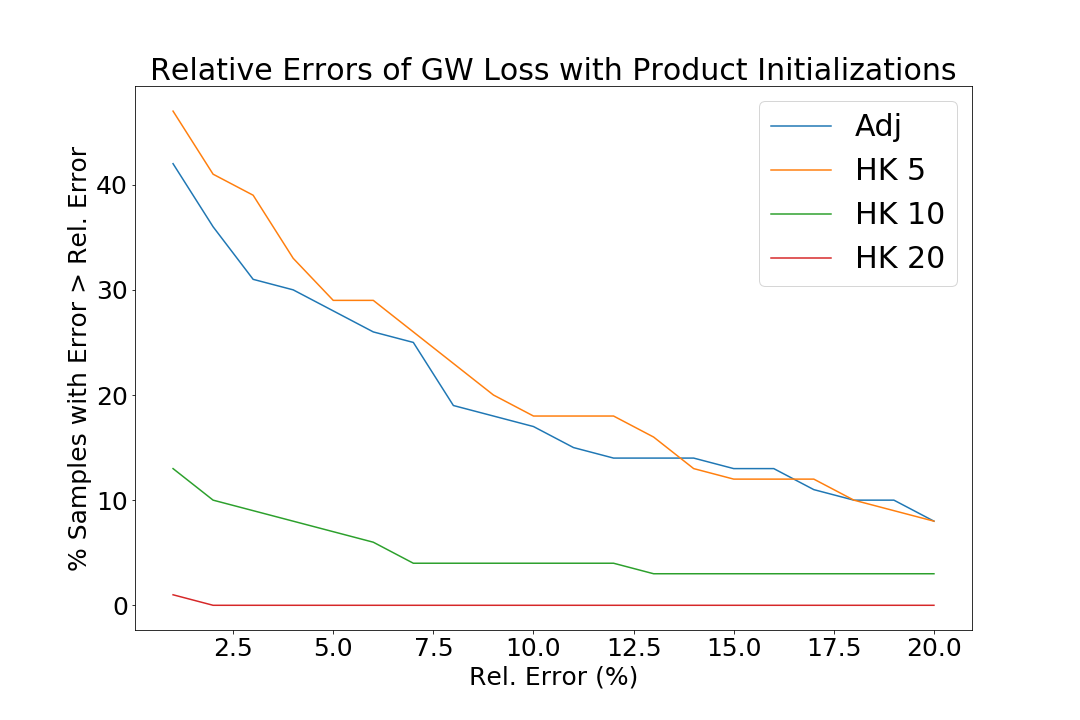}
\caption{Results of energy landscape experiments.}\label{fig:error_rates}
\end{figure}

Figure \ref{fig:error_rates} gives a more detailed view of the results reported in Table \ref{tab:mean_stats_energy_landscapes}. For each plot, the $x$-axis is (Worst or Product) error percentage. The $y$-axis shows the percentage of samples whose error was above the relative error rate. We see that a significant number of samples have high error rates for adjacency loss \eqref{eqn:gwl_loss} and spectral loss \eqref{eqn:gwspec_loss} with $t=5$. For spectral loss with $t=10$ or $20$, these error rates are greatly decreased. In particular, spectral loss with $t=20$ has essentially zero samples with error rate above 2\%.

\subsection{Additional figures}

\begin{figure}
    \centering
    \includegraphics[width=0.9\linewidth]{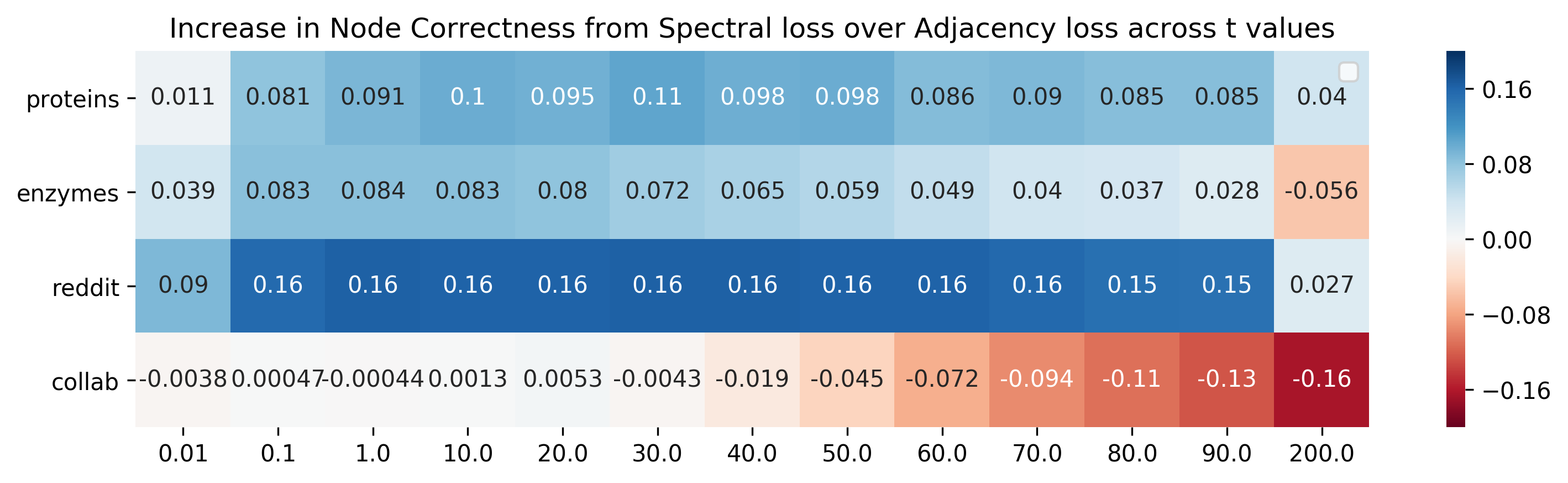}
    \caption{Improvement (blue cells) obtained by using spectral loss \eqref{eqn:gwspec_loss} instead of adjacency loss \eqref{eqn:gwl_loss} across a range of $t$ values ($x$ axis).}
    \label{fig:matching_scores_scale}
\end{figure}

Here we add some figures that help better understand some of the quantitative results presented in the main text. Figure \ref{fig:matching_scores_scale} shows that the improvement in the graph matching experiment obtained via SpecGWL remains stable across a wide range of scale parameters. Specifically, we computed SpecGWL loss for $t\in \{10,20,\ldots,90\}$ for each of the four datasets. The \textbf{Collab} dataset was the only one where there was no appreciable improvement from using SpecGWL, but there was no significant decrease in performance either for scales in the range $t \in \{10,20,30\}$.

\begin{figure}
    \centering
    \includegraphics[width=\linewidth]{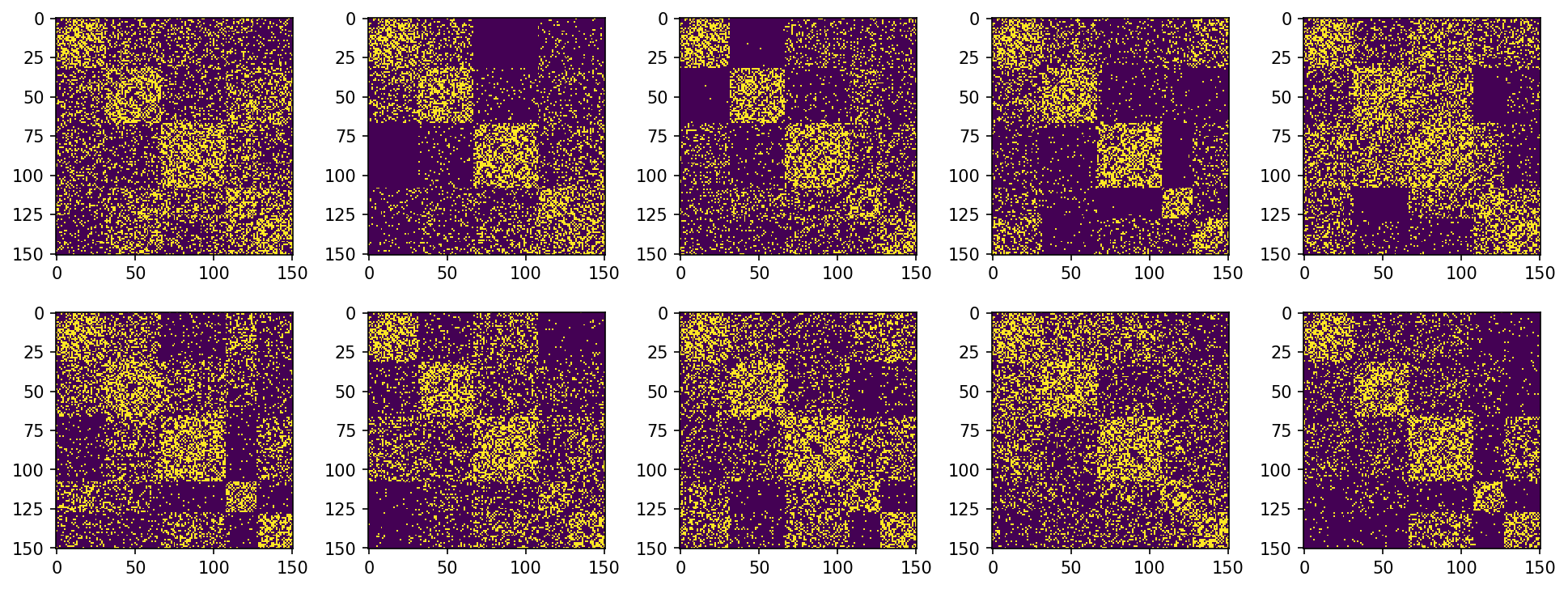}
    \caption{Stochastic block models used in the supervised partitioning task with cross-validation}
    \label{fig:partition-sbm-five}
\end{figure}

Figure \ref{fig:partition-sbm-five} shows the 10 stochastic block model networks used in the synthetic graph partitioning experiment. Each network has 5 blocks, and the block sizes were chosen uniformly at random from the range $[20,50]$ at the beginning of the experiment. Within-block edge densities were fixed at 0.5, and across-block edge densities were chosen uniformly at random in the range $[0,0.3]$. 

\subsection{Visualizing Graph Matchings}

Here we describe how the interpolations used to visualize coupling quality in Figure \ref{fig:matchings} were produced. Let $(G,p)$ and $(H,q)$ be measure graphs and $C \in \mathcal{C}(p,q)$ a coupling. To produce an interpolation, we first ``blow up'' $C$ so that it has the form of a weighted permutation matrix. This is done by first scanning across rows; any row with more than a single nonzero entry is split into ``dummy'' copies, each of which contains a single nonzero entry from the original row. The splits allow us to split nodes of $G$ into dummy copies, with weights given by entries in the corresponding row of $C$. The same procedure is applied to split columns of $C$ and to split nodes of $H$. The result is a pair of expanded measure graphs $(G',p')$ and $(H',q')$ together with an expanded coupling $C'$ which provides a bijective correspondence between the nodes of $G'$ and $H'$. Once such a bijective correspondence is obtained, we position each graph $G'$ and $H'$ in the plane using a common embedding modality and then performing Procrustes alignment of the resulting embeddings. To interpolate the graphs, we simply interpolate positions of the bijectively matched nodes, while phasing in new edges that are formed. This visualization method has strong theoretical justification: building on work of \cite{sturm2012space}, it is shown by \cite{gwa} that this process represents a geodesic (in the metric geometry sense) in the space of edge-weighted measure graphs. We observe that the conclusion of Lemma 1 is useful here, since the theoretical guarantee on the sparsity of $C$ implies that $C$ will not get too large in the ``blow up'' phase of the algorithm. 

To produce each example in Figure \ref{fig:matchings}, we sampled 100 couplings from the coupling polytope via the MCMC algorithm (1000 MCMC steps between each coupling) as initializations. We then computed an optimal coupling between the graphs by optimizing the relevant loss function from each initialization and keeping the coupling with the lowest loss from the resulting ensemble. 

\subsection{Dependence of Matchings on $t$-Value}

\begin{figure}
    \centering
    \includegraphics[width=0.8\linewidth]{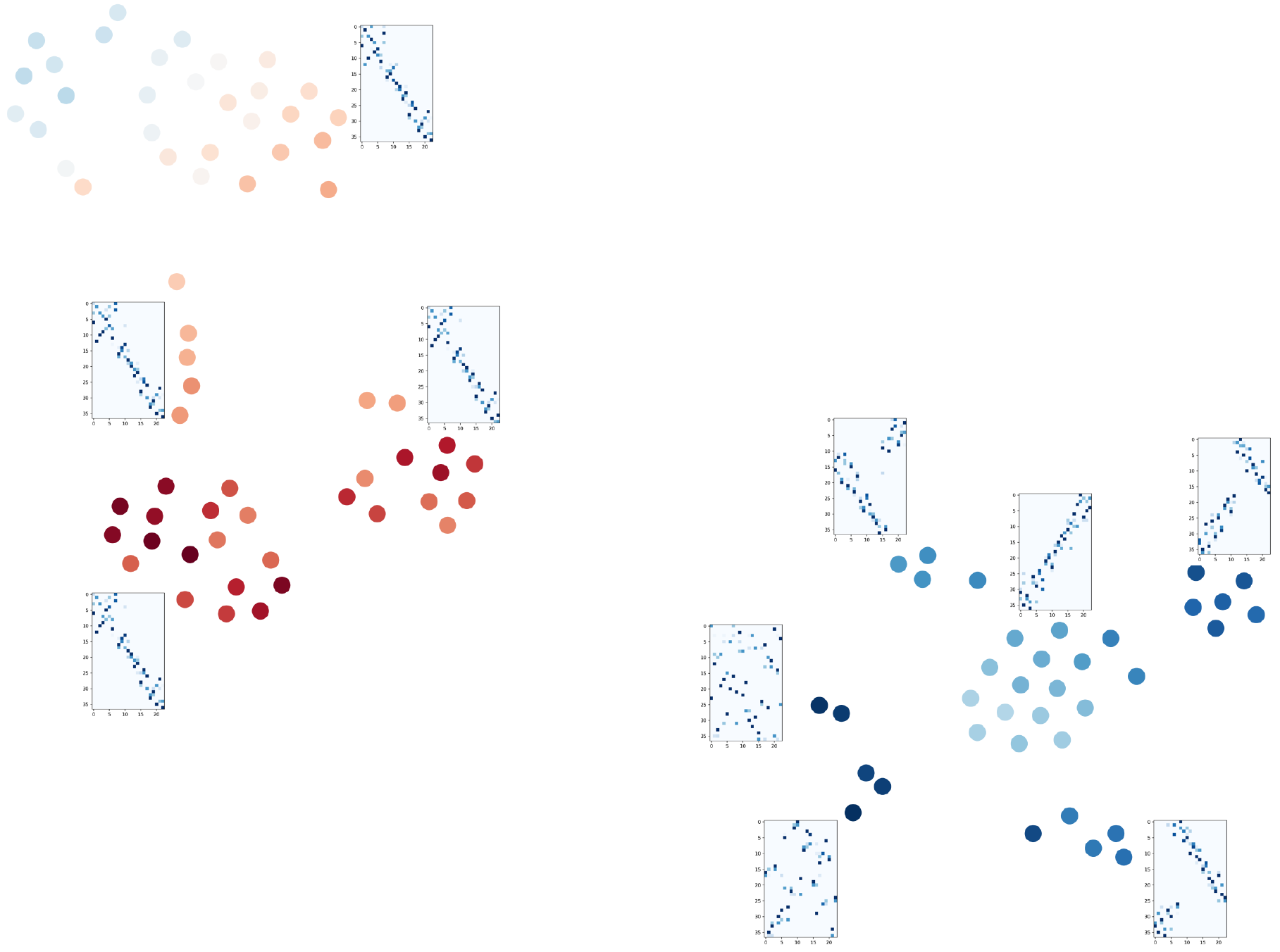}
    \caption{t-SNE embedding of optimal couplings between two graphs from the \textbf{Enzymes} dataset obtained via spectral loss \eqref{eqn:gwspec_loss} using 100 equally spaced $t$ values in the range $[0,50]$. The ground metric is the $l^2$ norm between vectorized representations of the couplings. The range $[0,50]$ is mapped linearly from the blue to red color scheme.}
    \label{fig:hkscale_tsne_composite}
\end{figure}

\begin{figure}
    \centering
    \includegraphics[width=0.8\linewidth]{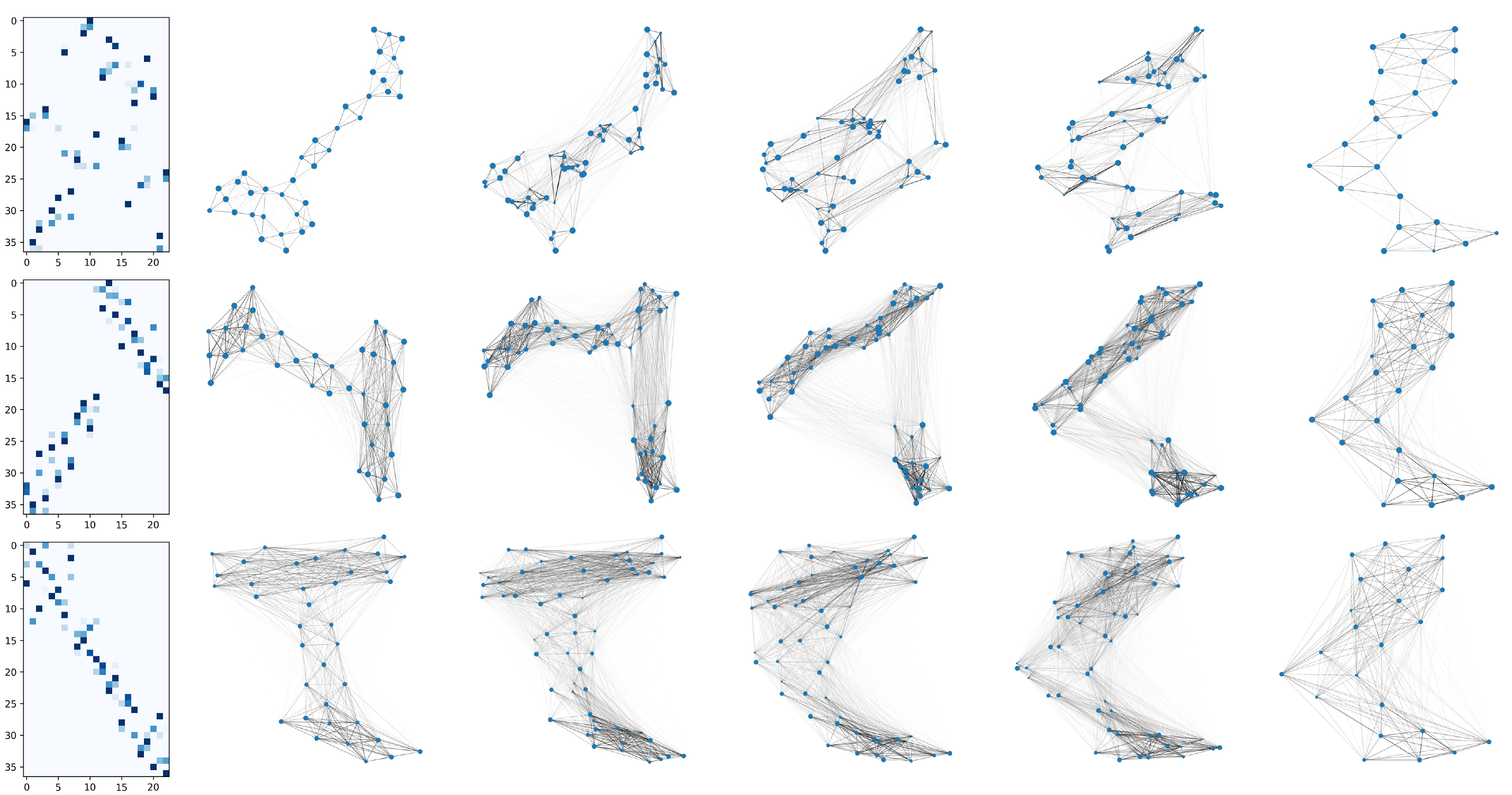}
    \caption{Interpolation visualizations obtained via some of the representative couplings in Figure \ref{fig:hkscale_tsne_composite} arranged top to bottom with increasing $t$. Note that the interpolation in the third row, corresponding to the largest $t$ value, represents global structure more faithfully.}
    \label{fig:hkscale_geodesic_composite}
\end{figure}

To understand the landscape of optimal couplings that occur for different scale parameters, we took two graphs from the \textbf{Enzymes} dataset and computed optimal couplings for spectral loss using 100 linearly spaced $t$ values in the range $[0,50]$. Figure \ref{fig:hkscale_tsne_composite} shows a t-SNE embedding (with perplexity = 15) obtained after flattening these couplings into vectors in Euclidean space. The inset coupling matrices are representatives of the points in each significant cluster. Interpolation visualizations for some of the coupling matrices from different clusters are provided in Figure \ref{fig:hkscale_geodesic_composite}.

\subsection{Averaging}

We use the observations regarding the energy landscape and the quality of matchings to show that in the \emph{GW averaging} problem, using the heat kernel leads to 10x faster convergence than the adjacency matrix, and moreover, the heat kernel yields a more ``unique'' barycenter. Specifically, given measure network representations $X_1,X_2,\ldots, X_n$, a \frechet mean is an element of $\argmin_X \sum_i\dgw(X,X_i)^2$. The objective of the GW averaging problem is to compute this barycenter, i.e. an average representation. In the Python OT package \citep{pot}, this barycenter is computed iteratively from a random initialization (cf. the \texttt{gromov\_barycenter} function). As a proxy for the ``uniqueness'' of the barycenter, we compute the barycenter for multiple random initializations, and then take the variance of the distribution of \frechet losses achieved by the barycenters.  

\begin{figure}
    \centering
    \caption{\textbf{Left:} Differences in \frechet loss of the GW average across representations. \textbf{Center:} Mean-centered \frechet loss, indicating the greater variance and sensitivity to initialization for the adjacency representation. \textbf{Right:} Distribution of runtimes shows 10x speedup for the heat kernel.}
    \label{fig:gwa1}
    \includegraphics[width=0.3\textwidth]{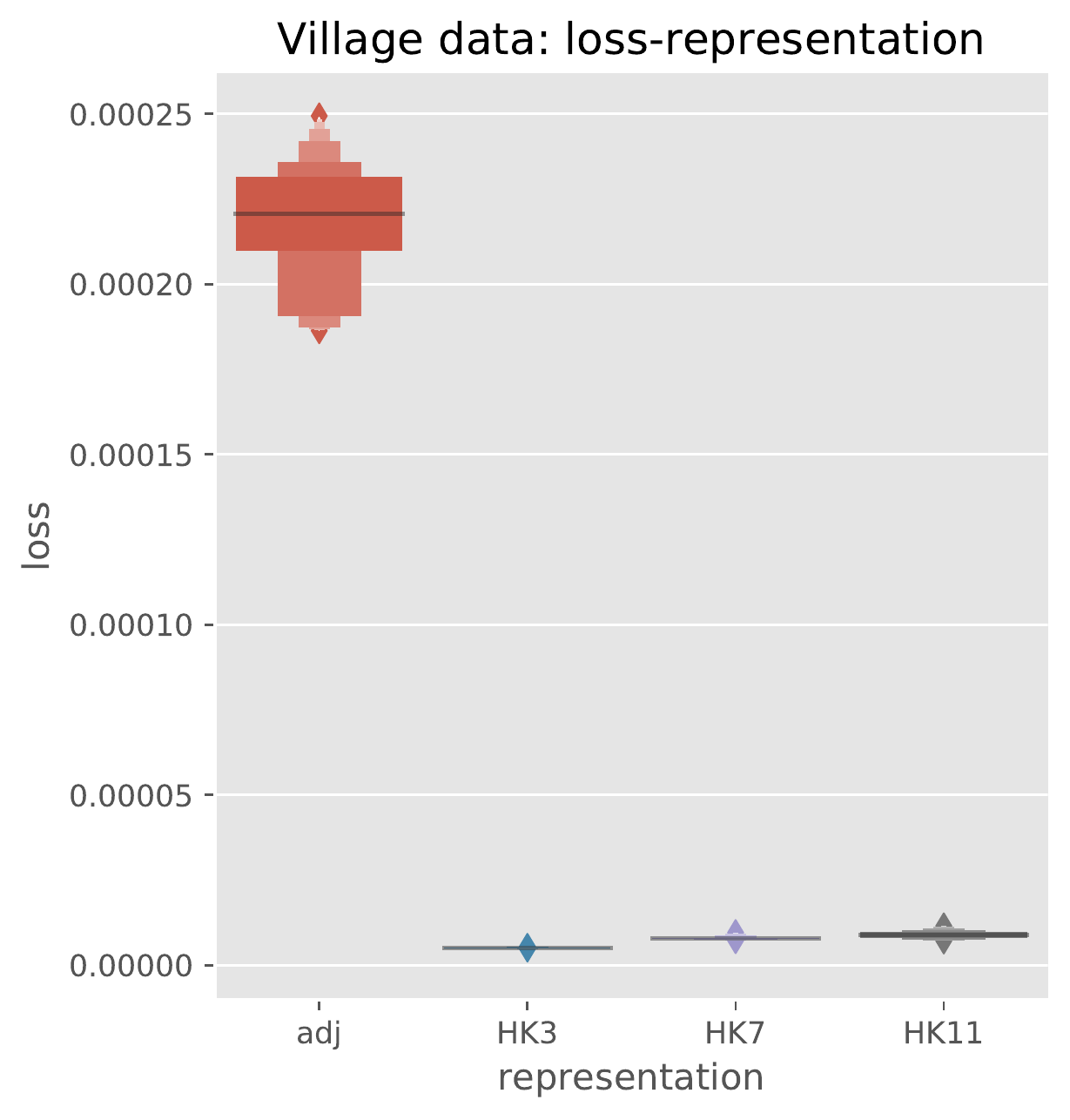}
    \includegraphics[width=0.315\textwidth]{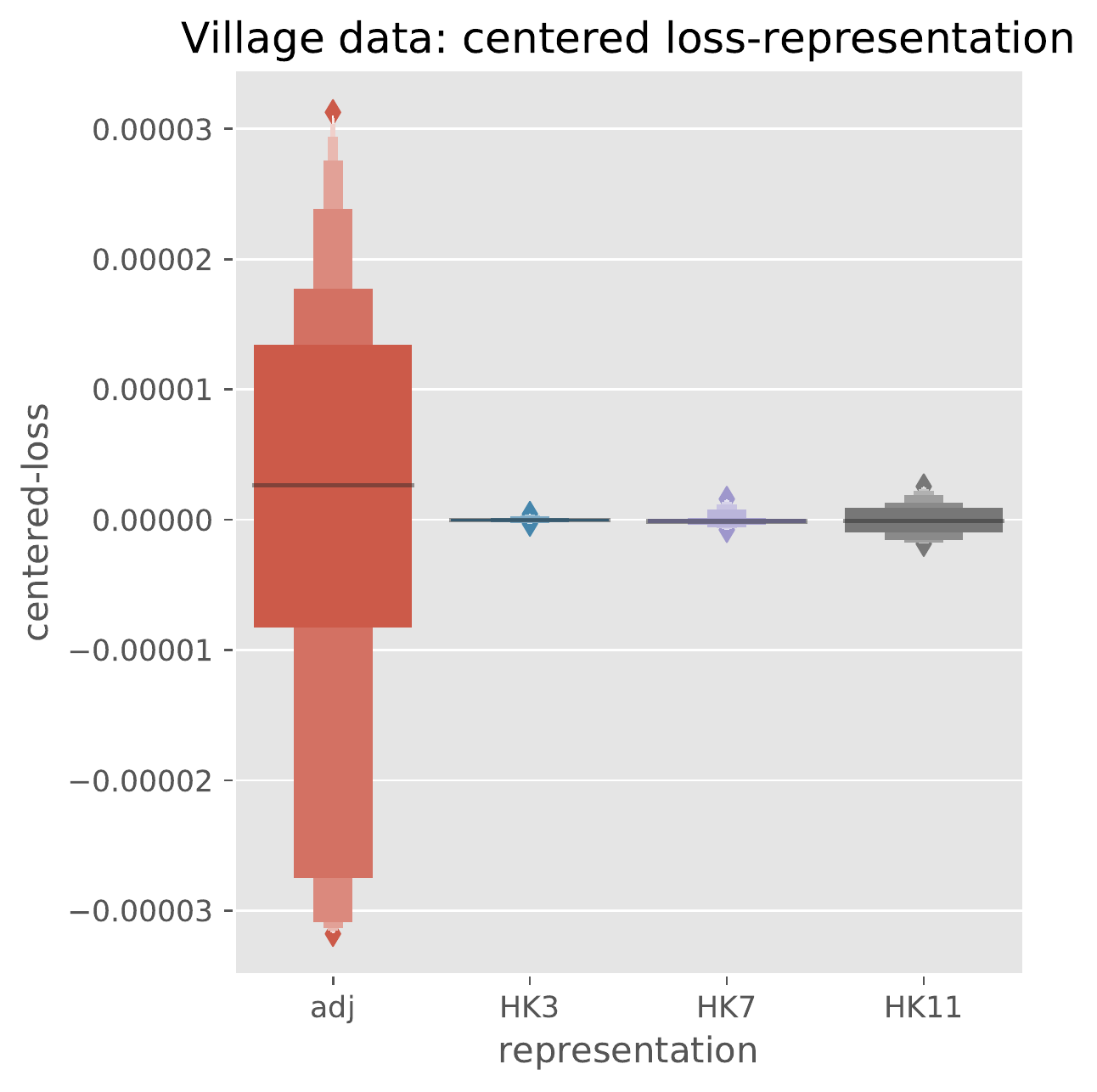}
    \includegraphics[width=0.3\textwidth]{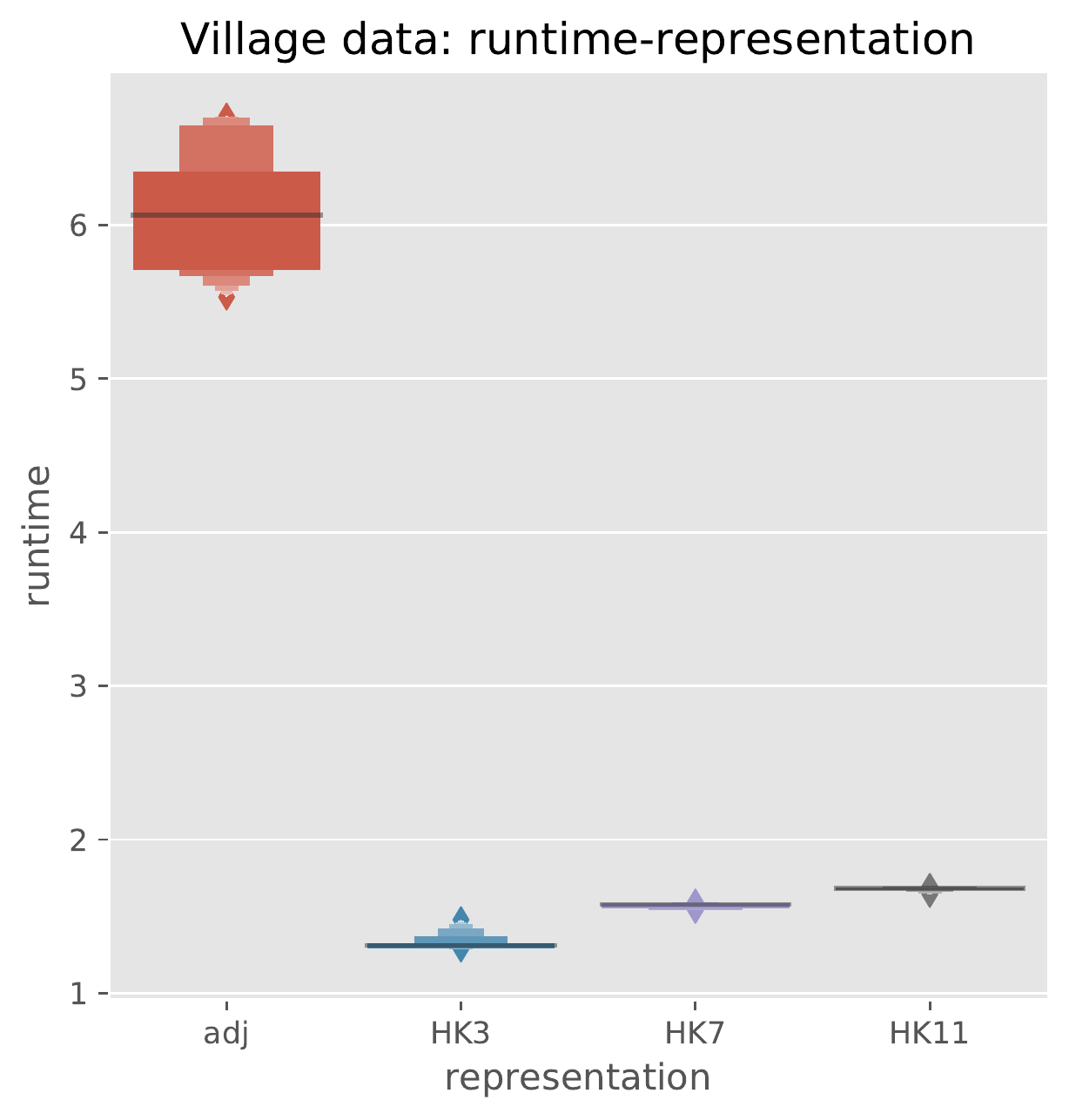}
\end{figure}

We demonstrate this claim on the \textbf{Village} dataset. We ran a bootstrapping procedure to sample 10 sets of 30 nodes, and took the induced subgraphs to obtain 10 subgraphs. To keep the samples from being too sparse, we first sorted the nodes in order of decreasing betweenness centrality, and then selected 30 nodes (for each iteration) from the top 40 nodes with the highest centrality. Next we computed both adjacency and heat kernel representations (for $t = 3,7,11$) of these subgraphs. Then we used the \texttt{gromov\_barycenter} function to compute averages of the adjacency and heat kernel representations. Each call to \texttt{gromov\_barycenter} uses a random initialization. Using this randomness as a source of stochasticity, we repeated the set of barycenter computations 10 times to obtain four distribution of \frechet losses. After mean-centering the distributions, the variance of the adjacency distribution was found to be two orders of magnitude higher than any of the heat kernel distributions, and each of the three comparisons was found to be statistically significant by computing Bartlett tests for unequal variance ($p<10^{-6}$ for all, adjusted for multiple comparison via Bonferroni correction). Boxplots of the results are shown in Figure \ref{fig:gwa1}.

\subsection{Graph Partitioning}

Runtimes for GWL and \gwspec on the graph partitioning experiment are reported in Table \ref{tab:avetimes}. For \gwspec\!, the times are averaged over several values of $t$, with the idea that finding the correct $t$-value is a preprocessing hyperparameter tuning step. For both GWL and \gwspec\!, partitionings were obtained using standard projected gradient descent. Speedups are obtained for GWL via the regularized proximal gradient method, but we were not able to obtain results on all datasets with this method due to numerical issues (see below). Runtimes for this method are also reported as GWL-Prox. We observe that spectral loss provides up to 10x acceleration in convergence rate for standard gradient descent and even outperforms the proximal gradient in compute time.

\begin{table}
    \centering
    \caption{Comparison between runtime of GWL and average runtime of \gwspec across $t$ parameters. ``-Prox'' rows use regularized proximal gradient with Sinkhorn iterations as used by \cite{xu2019scalable}. Other rows use vanilla gradient descent.}
    \label{tab:avetimes}
    \begin{tabular}{
    p{0.15\textwidth} p{0.03\textwidth} 
    p{0.04\textwidth} p{0.03\textwidth} p{0.04\textwidth} p{0.03\textwidth}
    p{0.04\textwidth} p{0.03\textwidth} p{0.04\textwidth} p{0.03\textwidth} p{0.04\textwidth} p{0.03\textwidth} p{0.04\textwidth} 
    }\toprule
    \multirow[t]{3}{*}{Method} & 
    \multicolumn{4}{c}{Wikipedia} &
    \multicolumn{4}{c}{EU-email} & 
    \multicolumn{2}{c}{Amazon} &
    \multicolumn{2}{c}{Village}\\
    \cmidrule(r){2-5} \cmidrule(r){6-9}
    \cmidrule(r){10-11}
    \cmidrule(r){12-13}
    {} & \multicolumn{2}{c}{sym} &
    \multicolumn{2}{c}{asym} & 
    \multicolumn{2}{c}{sym} &
    \multicolumn{2}{c}{asym} &
    {} & {} & {} & {} \\
    \cmidrule(r){2-3} \cmidrule(r){4-5} 
    \cmidrule(r){6-7}
    \cmidrule(r){8-9}
    {} & raw & noisy & raw & noisy
    & raw & noisy & raw & noisy
    & raw & noisy & raw & noisy\\
    \midrule
    GWL & 14.1 & 16.1 & 16.0 & 14.2 & 1.5 & 6.7 & \textbf{0.9} & 1.6 & 8.6 & 13.1 & 3.3 & 5.4\\
    \gwspec & \textbf{1.8} & \textbf{2.3} & \textbf{2.4} & \textbf{2.4} & \textbf{1.0} & \textbf{0.9} & \textbf{0.9} & \textbf{1.0} & \textbf{1.4} & \textbf{0.9} & \textbf{1.8} & \textbf{2.7}\\
    \midrule
    GWL-Prox & --- & --- & --- & --- & \textbf{0.9} & \textbf{0.8} & --- & --- & \textbf{1.2} & \textbf{1.0} & 2.2 & 2.2 \\
    \multirow[t]{2}{*}{SpecGWL-Prox} & 2.9 & 2.6 & 2.9 & 2.9 & 
    \textbf{0.9} & 1.0 & 1.0 & 1.0 & 1.3 & 1.8 & \textbf{1.8} & \textbf{2.0}\\ 
    \bottomrule
    \end{tabular}
    
\end{table}

When employing the regularized proximal gradient method, we found that the results were sensitive to the choice of regularization parameter $\beta$ (as is also observed by \cite{xu2019scalable}), leading to numerical blowups if not chosen carefully. In reporting each of the results below, we hand-tuned $\beta$ after testing in the $10^{-1}, 10^{-2}, 10^{-3},\ldots, 10^{-9}$ regimes. For \textbf{Wikipedia}, we used $\beta=2\cdot 10^{-5}$ for \gwspec\!, but were unable to find a $\beta$ that provided stable results for GWL. For \textbf{EU-email}, we used $2\cdot 10^{-7}$ for GWL and $3\cdot 10^{-8}$ for \gwspec\!. For \textbf{Amazon}, we used $\beta=4\cdot 10^{-3}$ for GWL and $1.5\cdot 10^{-6}$ \gwspec\!. Finally, for \textbf{Village} we used $\beta=5\cdot 10^{-6}$ for \gwspec\!. This $\beta$ led to numerical instability for GWL, but $\beta=5\cdot 10^{-5}$ worked and yielded the results we report below. In summary, it appears that the structure of the graph has a significant effect on the optimal choice of regularization parameter (e.g. the Wikipedia graph is relatively very sparse). Because the numerical instability issues are very sensitive to the regularization, one avenue for future work could be to incorporate the strategies described in the PhD thesis of  \cite{chizat-thesis} (e.g. ``absorption into the log domain'') to stabilize the regularized proximal gradient method.

\begin{table}
    \centering
    \caption{Performance of GWL and \gwspec using regularized proximal gradient descent and Sinkhorn iterations as in \cite{xu2019scalable}. '---' denotes that an AMI score could not be calculated due to numerical instability.}
    \label{tab:partition-sinkhorn}
    \begin{tabular}{
    p{0.15\textwidth} p{0.03\textwidth} 
    p{0.04\textwidth} p{0.03\textwidth} p{0.04\textwidth} p{0.03\textwidth}
    p{0.04\textwidth} p{0.03\textwidth} p{0.04\textwidth} p{0.03\textwidth} p{0.04\textwidth} p{0.03\textwidth} p{0.04\textwidth} 
    }\toprule
    \multirow[t]{3}{*}{Method} & 
    \multicolumn{4}{c}{Wikipedia} &
    \multicolumn{4}{c}{EU-email} & 
    \multicolumn{2}{c}{Amazon} &
    \multicolumn{2}{c}{Village}\\
    \cmidrule(r){2-5} \cmidrule(r){6-9}
    \cmidrule(r){10-11}
    \cmidrule(r){12-13}
    {} &
    \multicolumn{2}{c}{asym} & 
    \multicolumn{2}{c}{sym} &
    \multicolumn{2}{c}{asym} &
    {} & {} & {} & {} \\
    \cmidrule(r){2-3} \cmidrule(r){4-5} 
    \cmidrule(r){6-7}
    \cmidrule(r){8-9}
    {} & raw & noisy & raw & noisy
    & raw & noisy & raw & noisy
    & raw & noisy & raw & noisy\\
    \midrule
    GWL-Prox & --- & --- & --- & --- & \textbf{0.45} & \textbf{0.40} & --- & --- & 0.49 & 0.39 & 0.72* & 0.58\\
    \multirow[t]{2}{*}{SpecGWL-Prox} & 0.51 & 0.39 & 0.39 & 0.29 & 0.01 & 0.01 & 0.03 & 0.03 & \textbf{0.66} & \textbf{0.43} & \textbf{0.84} & \textbf{0.72}\\
    \bottomrule
    \end{tabular}\\
    {\small *The code provided by \cite{xu2019scalable} included a representation matrix as \texttt{database[`cost']}, and this yielded the score of 0.72. However, this matrix was asymmetric and not equal to the symmetrized adjacency matrix that was used in experiments with other benchmarks. When using a symmetrized adjacency matrix, the score drops from 0.72 to 0.66.} 
\end{table}

\end{document}